\documentclass[10pt,twocolumn,letterpaper]{article}

\usepackage{iccv}
\usepackage{times}
\usepackage{epsfig}
\usepackage{graphicx}
\usepackage{amsmath}
\usepackage{amssymb}

\usepackage{caption}

\usepackage{makecell}
\usepackage{booktabs}
\usepackage{multirow}
\usepackage{siunitx}

\usepackage[T1]{fontenc}
\usepackage{fix-cm}

\newcolumntype{?}{!{\vrule width 1pt}}
\newcolumntype{C}[1]{>{\centering}m{#1}}

\newcolumntype{X}{@{\hskip\tabcolsep\vrule width 1.5pt\hskip\tabcolsep}}

\newcommand{\myfigurefivecol}[1]{
\begin{minipage}[b]{.17\textwidth}
\includegraphics[width=1.075\linewidth]{#1}
\end{minipage}
}

\newcommand{\myfigurethreecol}[1]{
\begin{minipage}[b]{.14\textwidth}
\includegraphics[width=1.1\linewidth]{#1}
\end{minipage}
}

\usepackage[pagebackref=true,breaklinks=true,letterpaper=true,colorlinks,bookmarks=false]{hyperref}

 \iccvfinalcopy 

\def\iccvPaperID{1306} 

\ificcvfinal\fi
\begin{document}

\title{High-for-Low and Low-for-High:\\Efficient Boundary Detection from Deep Object Features\\and its Applications to High-Level Vision}

\author{Gedas Bertasius\\
University of Pennsylvania\\
{\tt\small gberta@seas.upenn.edu}
\and
Jianbo Shi\\
University of Pennsylvania\\
{\tt\small jshi@seas.upenn.edu}
\and
Lorenzo Torresani\\
Dartmouth College\\
{\tt\small lt@dartmouth.edu}
}

\newcommand{\HfL}{\textsc{HfL}\xspace}

\maketitle

\def\GB#1{{{{\color{red}  #1}}}}
\def\LT#1{{{{\color{green}  #1}}}}






\begin{abstract}

Most of the current boundary detection systems rely exclusively on low-level features, such as color and texture. However, perception studies suggest that humans employ object-level reasoning when judging if a particular pixel is a boundary. Inspired by this observation, in this work we show how to predict boundaries by exploiting object-level features from a pretrained object-classification network. Our method can be viewed as a \textit{High-for-Low} approach where high-level object features inform the low-level boundary detection process. Our model achieves state-of-the-art performance on an established boundary detection benchmark and it is efficient to run.

Additionally, we show that due to the semantic nature of our boundaries we can use them to aid a number of high-level vision tasks. We demonstrate that by using our boundaries we improve the performance of state-of-the-art methods on the problems of semantic boundary labeling, semantic segmentation and object proposal generation. We can view this process as a \textit{Low-for-High} scheme, where low-level boundaries aid high-level vision tasks. 

Thus, our contributions include a boundary detection system that is accurate, efficient, generalizes well to multiple datasets, and is also shown to improve existing state-of-the-art high-level vision methods on three distinct tasks.

\end{abstract}

\section{Introduction}

 \begin{table}[ht]
 \scriptsize
\begin{center}
\begin{tabular}{ c | c X c c | c | c |}
   \cline{2-6}
     & \multicolumn{1}{ c X}{Low-Level Task} & \multicolumn{4}{ c |}{ High-Level Tasks}\\
    \cline{2-6}
     & \multicolumn{1}{ c X}{BD} & \multicolumn{2}{ c |}{ SBL }& \multicolumn{1}{ c |}{SS} & \multicolumn{1}{ c |}{OP}\\  \cline{2-6}
     & ODS & MF & AP & PI-IOU & MR\\
    \hline
    \multicolumn{1}{| c |}{SotA} & 0.76~\cite{Shen_2015_CVPR}  & 28.0~\cite{BharathICCV2011} &19.9~\cite{BharathICCV2011} & 45.8~\cite{DBLP:journals/corr/ChenPKMY14} & 0.88~\cite{ZitnickDollarECCV14edgeBoxes}\\ \hline
    \multicolumn{1}{| c |}{\bf \HfL} & \bf 0.77 & \bf 62.5 & \bf 54.6 & \bf 48.8 & \bf 0.90\\ \hline
\end{tabular}
\end{center}
\caption{Summary of results achieved by our proposed method (\HfL) and state-of-the-art methods (SotA). We provide results on four vision tasks: Boundary Detection (BD), Semantic Boundary Labeling (SBL), Semantic Segmentation (SS), and Object Proposal (OP).  The evaluation metrics include ODS F-score for BD task, max F-score (MF) and average precision (AP) for SBL task, per image intersection over union (PI-IOU) for SS task, and max recall (MR) for OP task. Our method produces better results in each of these tasks according to these metrics.\vspace{-0.3cm}}
\label{summary}
\end{table}

In the vision community, boundary detection has always been considered a low-level problem. However, psychological studies suggest that when a human observer perceives boundaries, object level reasoning is used~\cite{psych,sanguinetti2013ground,KourtziKanwisher01}. Despite these findings, most of the boundary detection methods rely exclusively on low-level color and gradient features. In this work, we present a method that uses object-level features to detect boundaries. We argue that using object-level information to predict boundaries is more similar to how humans reason. Our boundary detection scheme can be viewed as a \textit{High-for-Low} approach where we use high-level object features as cues for a low-level boundary detection process. Throughout the rest of the paper, we refer to our proposed boundaries as \textit{High-for-Low} boundaries (\HfL).

We present an efficient deep network that uses object-level information to predict the boundaries. Our proposed architecture reuses features from the sixteen convolutional layers of the network of Simonyan et al.~\cite{Simonyan14c}, which we refer to as VGG net. The VGG net has been trained for object classification, and therefore, reusing its features allows our method to utilize high-level object information to predict \HfL boundaries. In the experimental section, we demonstrate that using object-level features produces semantically meaningful boundaries and also achieves above state-of-the-art boundary detection accuracy. 

Additionally, we demonstrate that we can successfully apply our \HfL boundaries to a number of high-level vision tasks. We show that by using \HfL boundaries we improve the results of three existing state-of-the-art methods on the tasks of semantic boundary labeling, semantic segmentation and object proposal generation. Therefore, using \HfL boundaries to boost the results in high level vision tasks can be viewed as a \textit{Low-for-High} scheme, where boundaries serve as low-level cues to aid high-level vision tasks. 

 We present the summarized results for the boundary detection and the three mentioned high-level vision tasks in Table~\ref{summary}. Specifically, we compare our proposed method and an appropriate state-of-the-art method for that task. As the results indicate, we achieve better results in each of the tasks for each presented evaluation metric. We present more detailed results for each of these tasks in the later sections.

In summary, our contributions are as follows. First, we show that using object-level features for boundary detection produces perceptually informative boundaries that outperform prior state-of-the-art boundary detection methods. Second, we demonstrate that we can use \HfL boundaries to enhance the performance on the high-level vision tasks of semantic boundary labeling, semantic segmentation and object proposal. Finally, our method can detect boundaries in near-real time. Thus, we present a boundary detection system that is accurate, efficient, and is also applicable to high level vision tasks.


\section{Related Work}

 \begin{figure}
\begin{center}
   \includegraphics[width=1\linewidth]{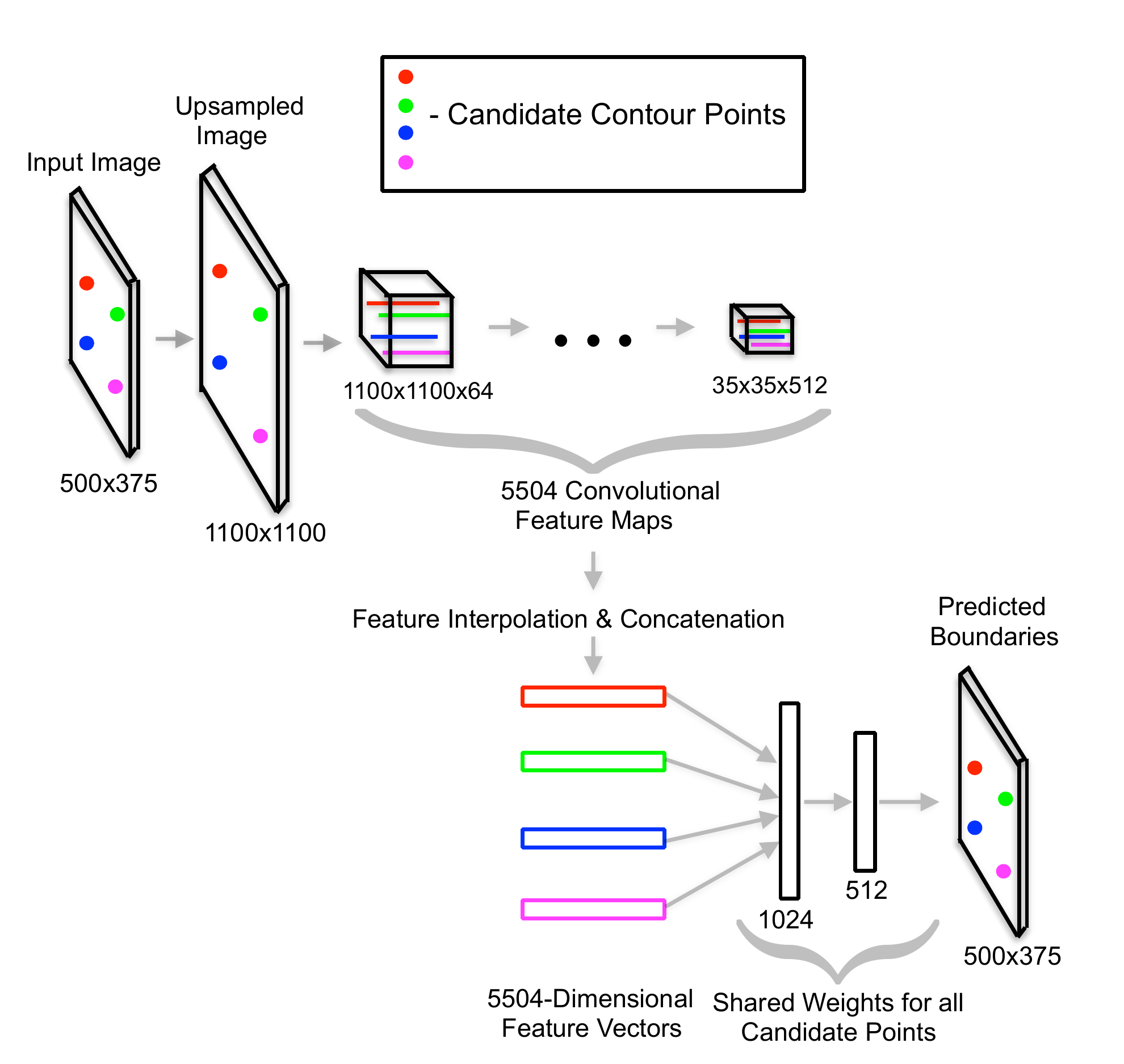}
\end{center}
   \caption{An illustration of our architecture (best viewed in color). First we extract a set of candidate contour points. Then we upsample the image and feed it through $16$ convolutional layers pretrained for object classification. For each candidate point, we find its correspondence in each of the feature maps and perform feature interpolation. This yields a $5504$-dimensional feature vector for each candidate point. We feed each of these vectors to two fully connected layers and store the predictions to produce a final boundary map.\vspace{-0.4cm}}
\label{fig:arch}
\end{figure}

   
\captionsetup{labelformat=empty}

\begin{figure*}
\centering

\myfigurefivecol{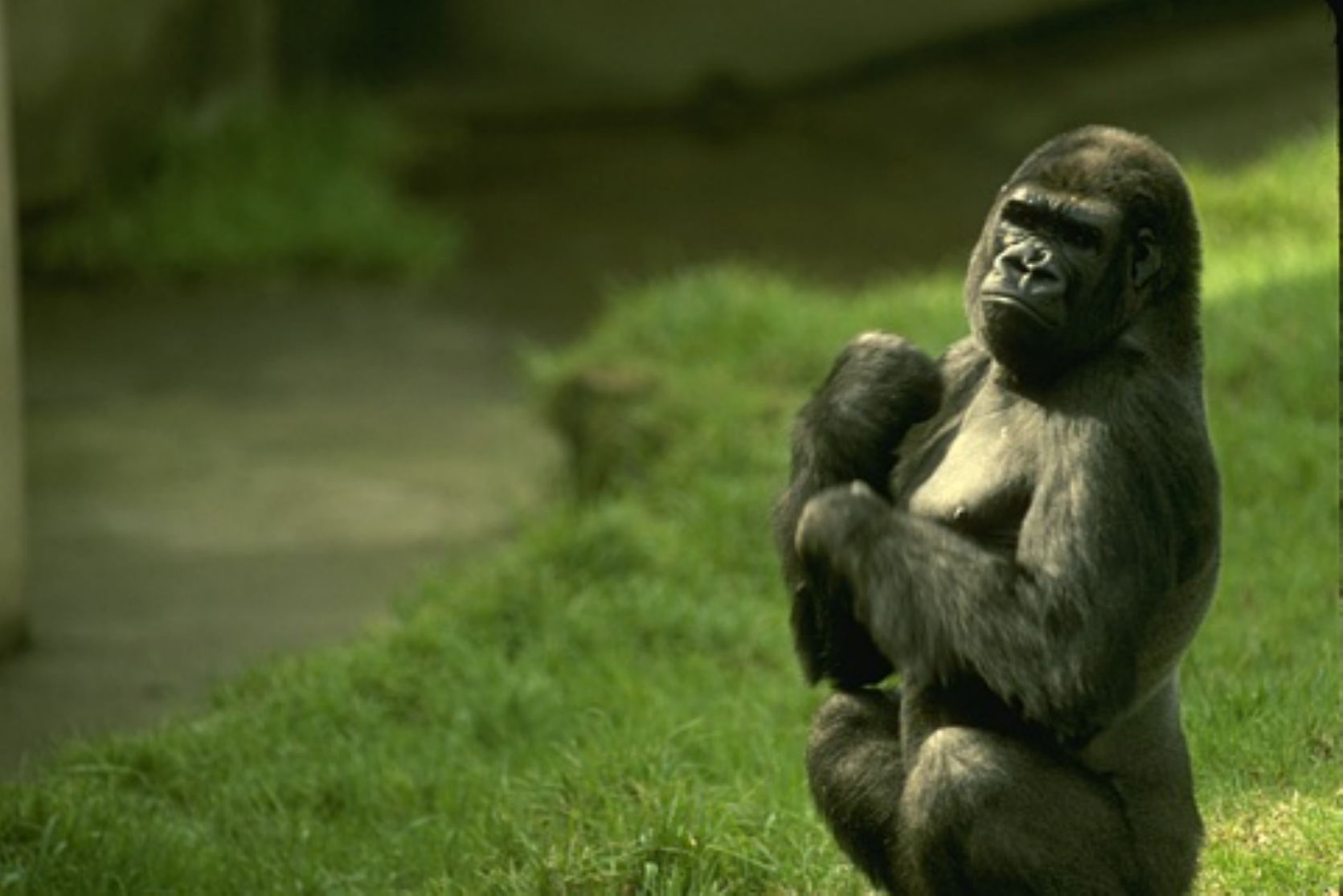}
\myfigurefivecol{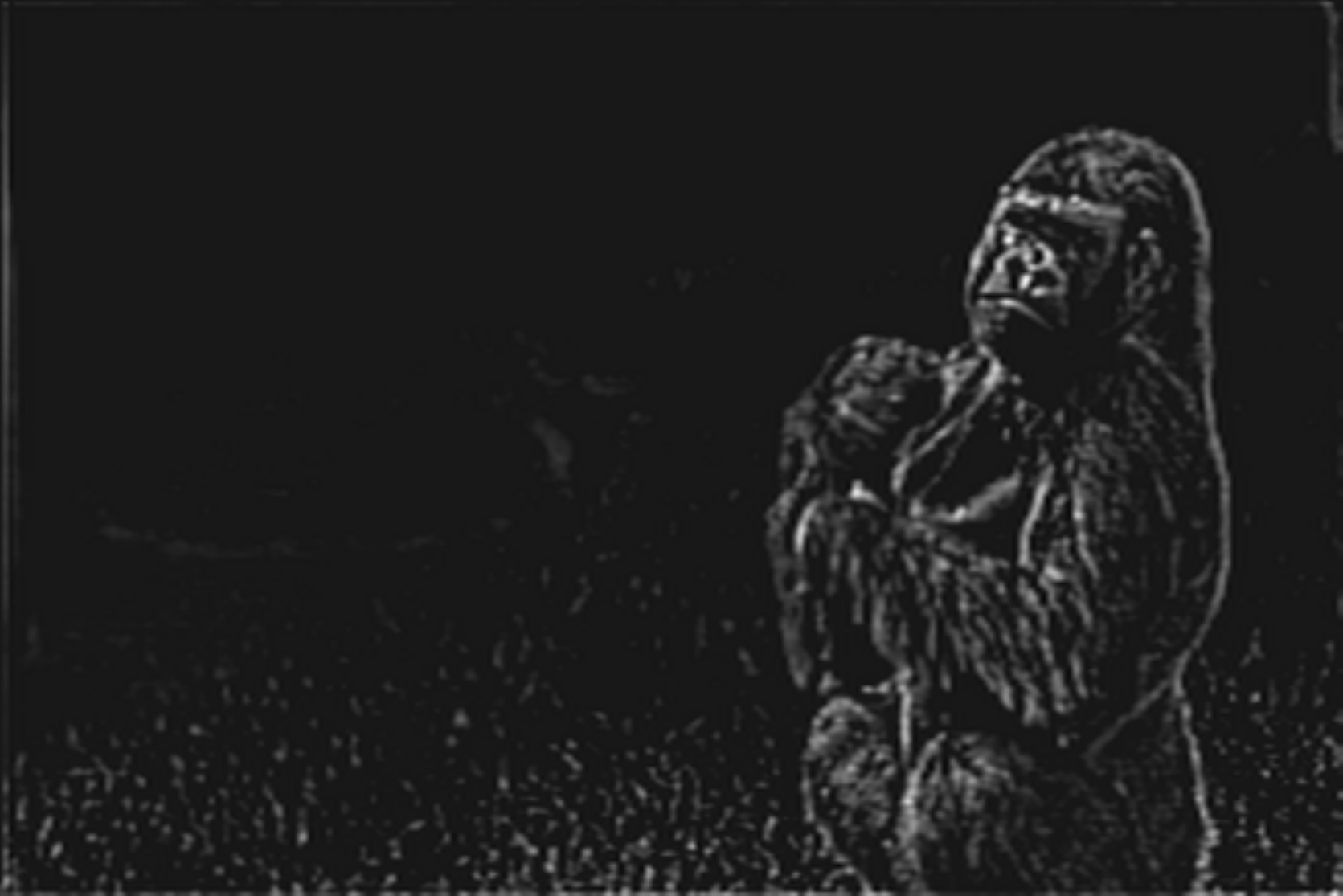}
\myfigurefivecol{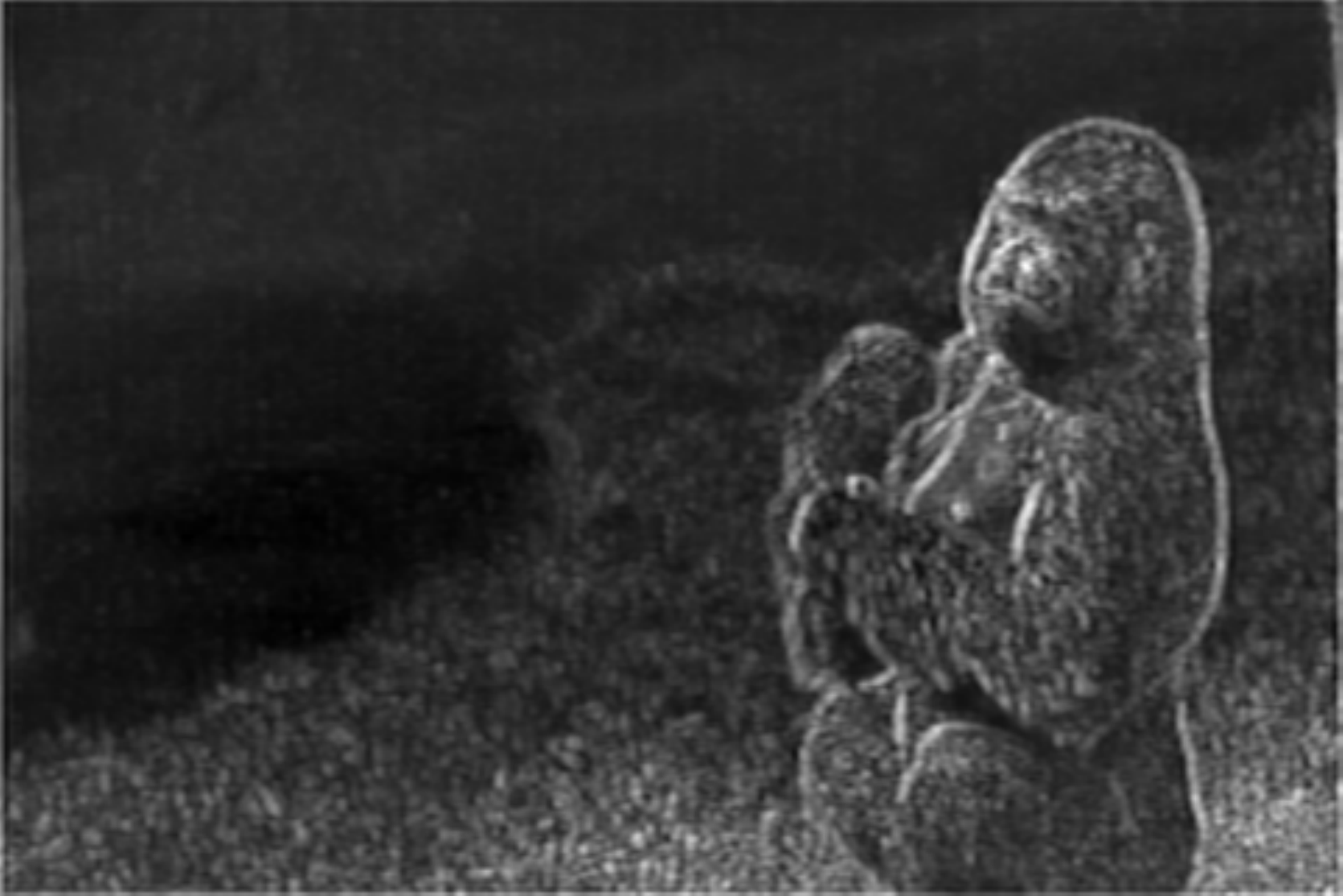}
\myfigurefivecol{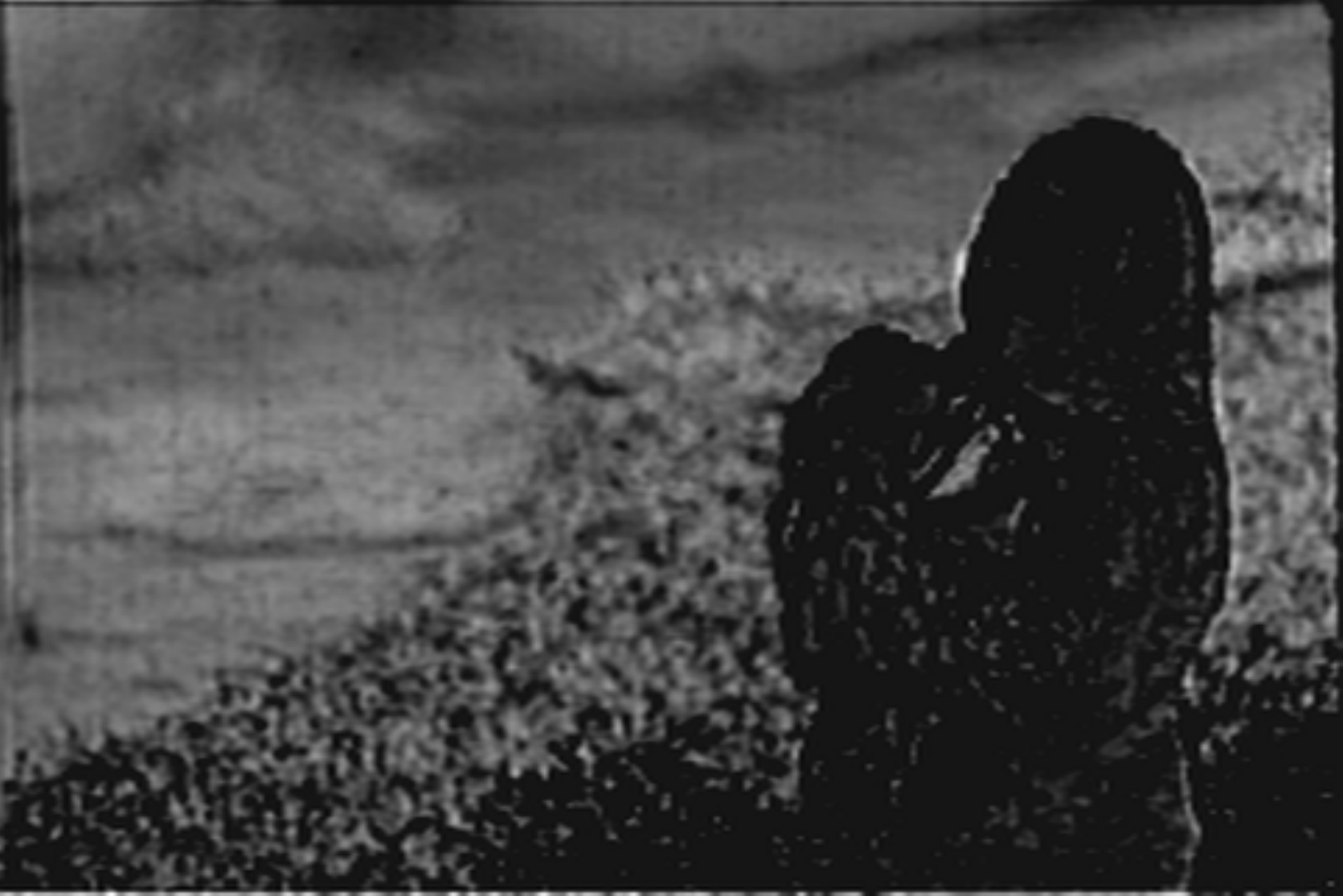}
\myfigurefivecol{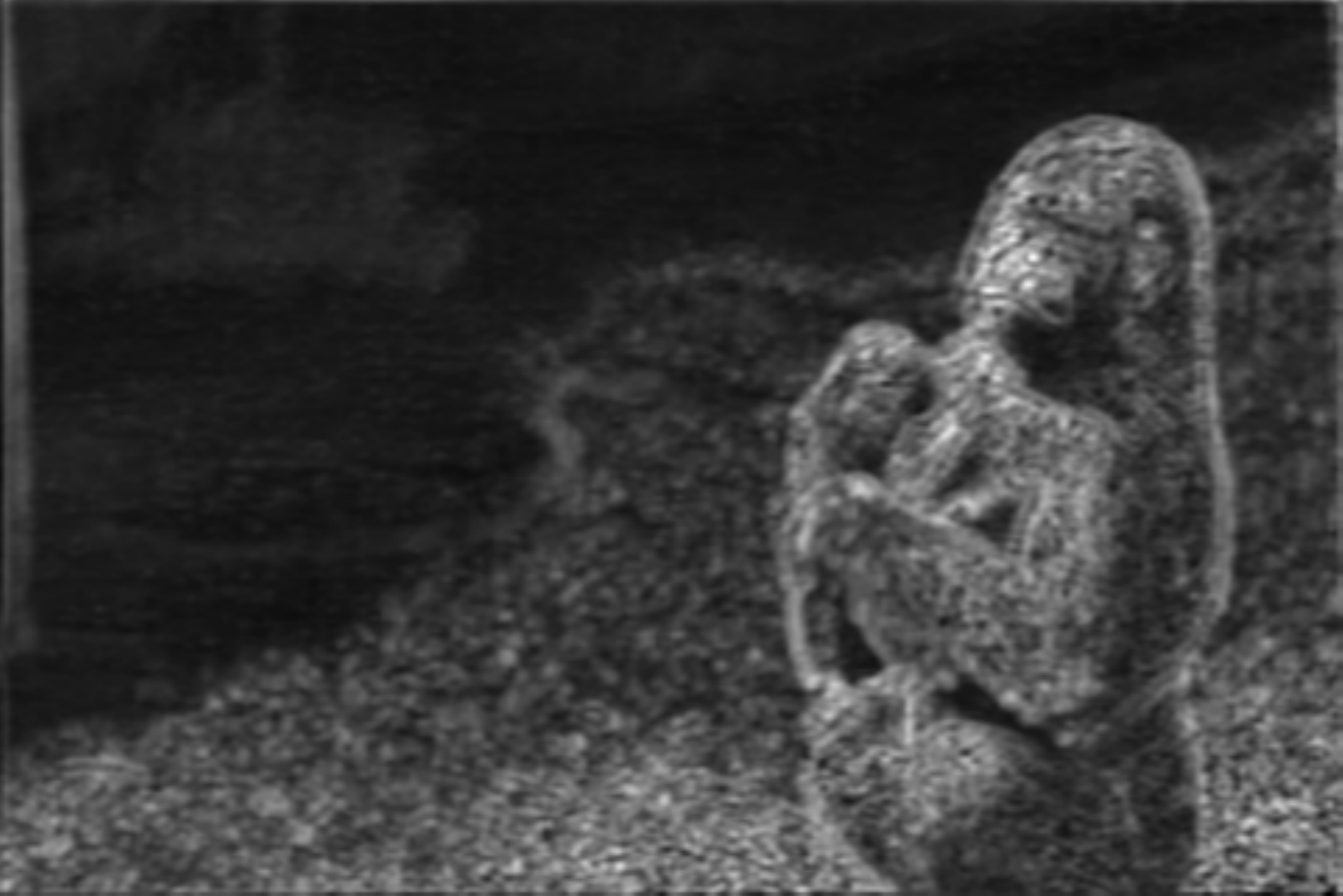}


\myfigurefivecol{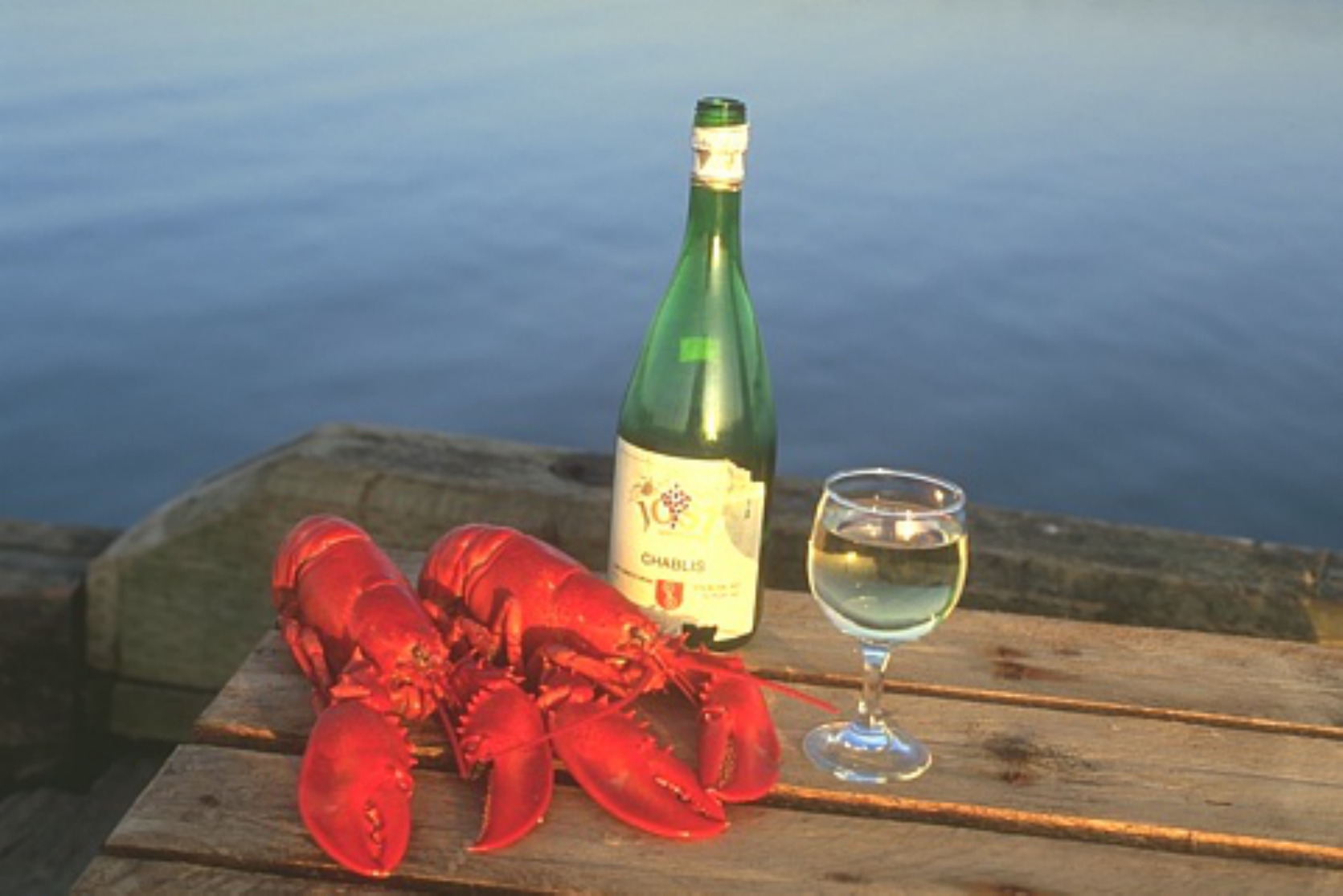}
\myfigurefivecol{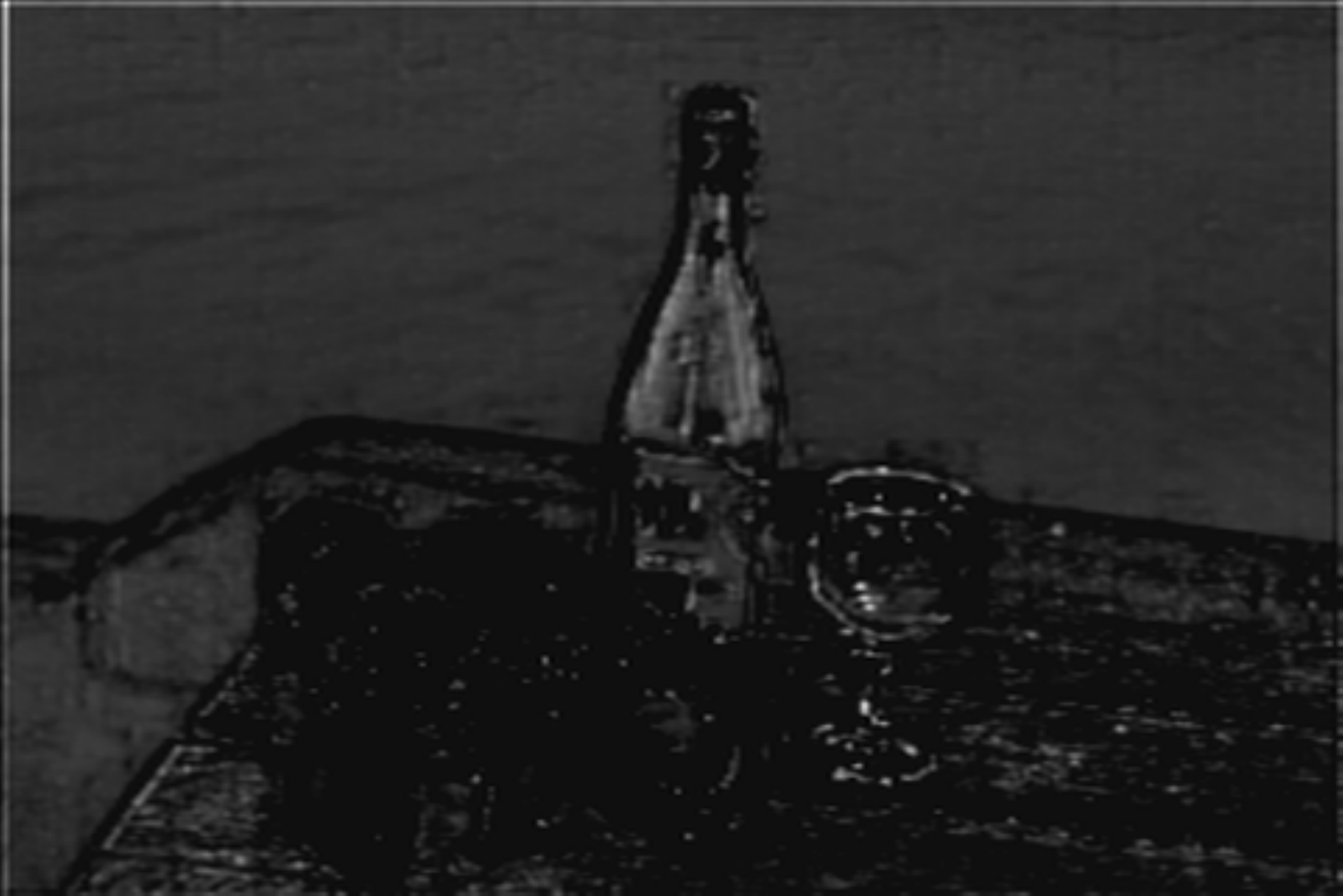}
\myfigurefivecol{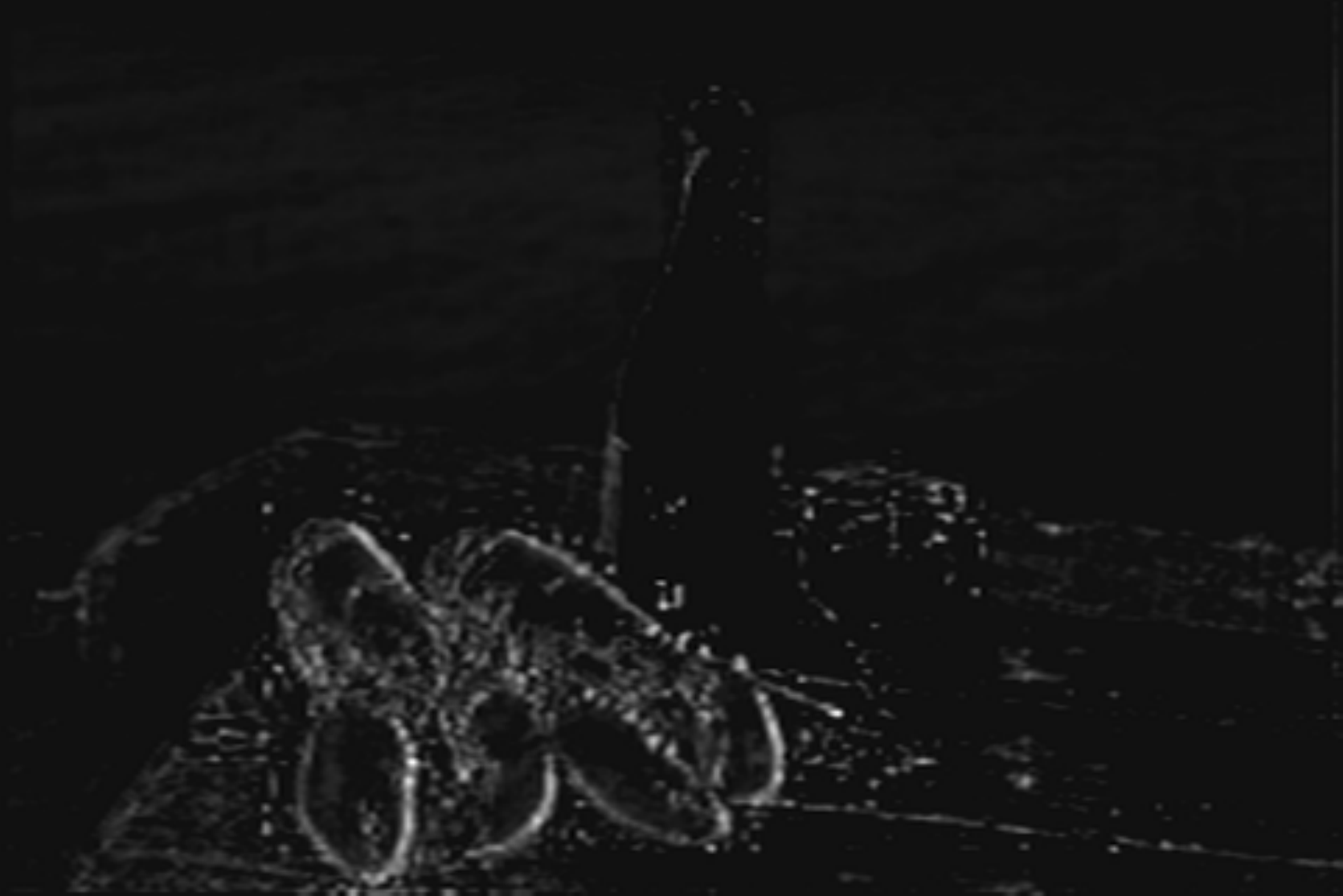}
\myfigurefivecol{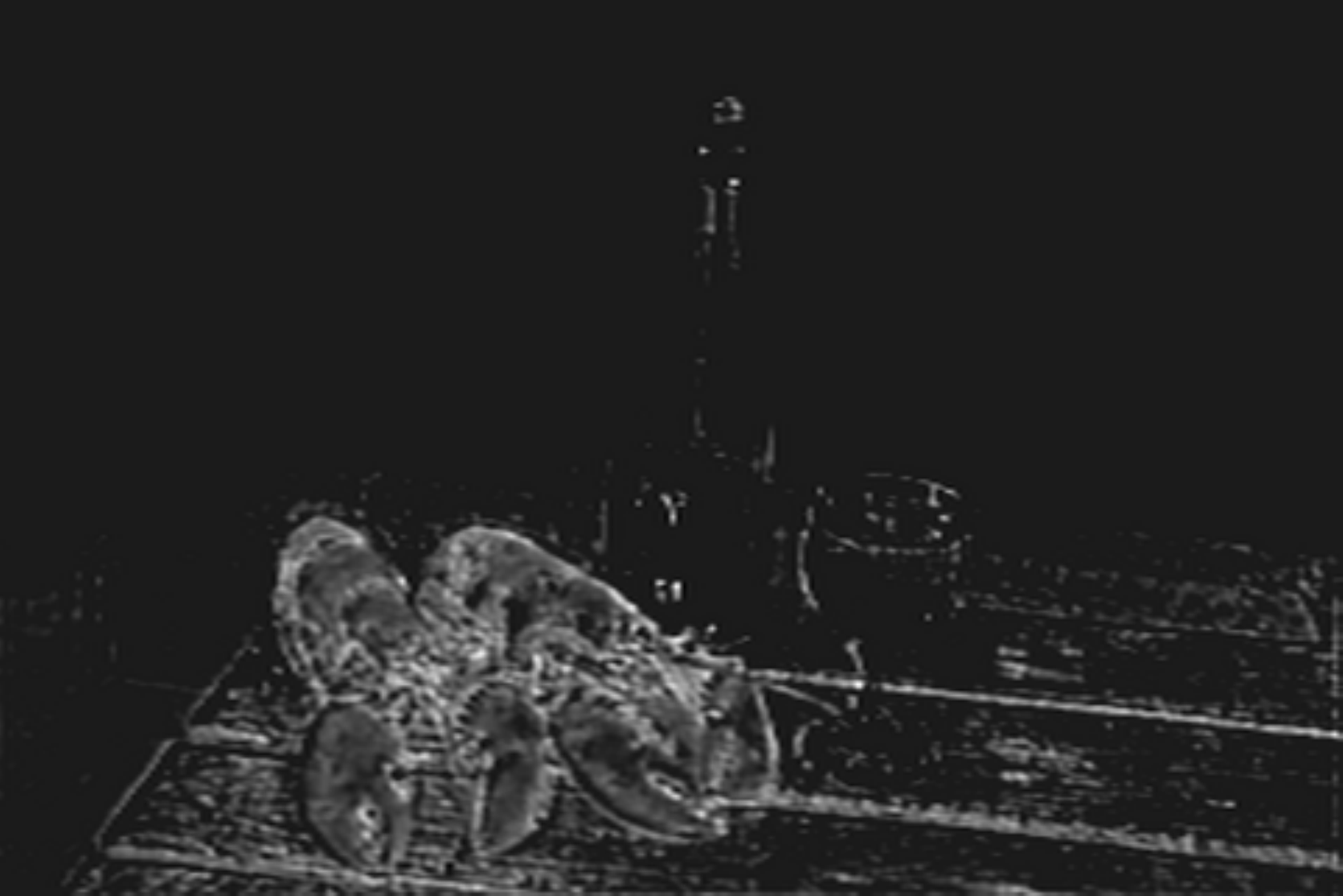}
\myfigurefivecol{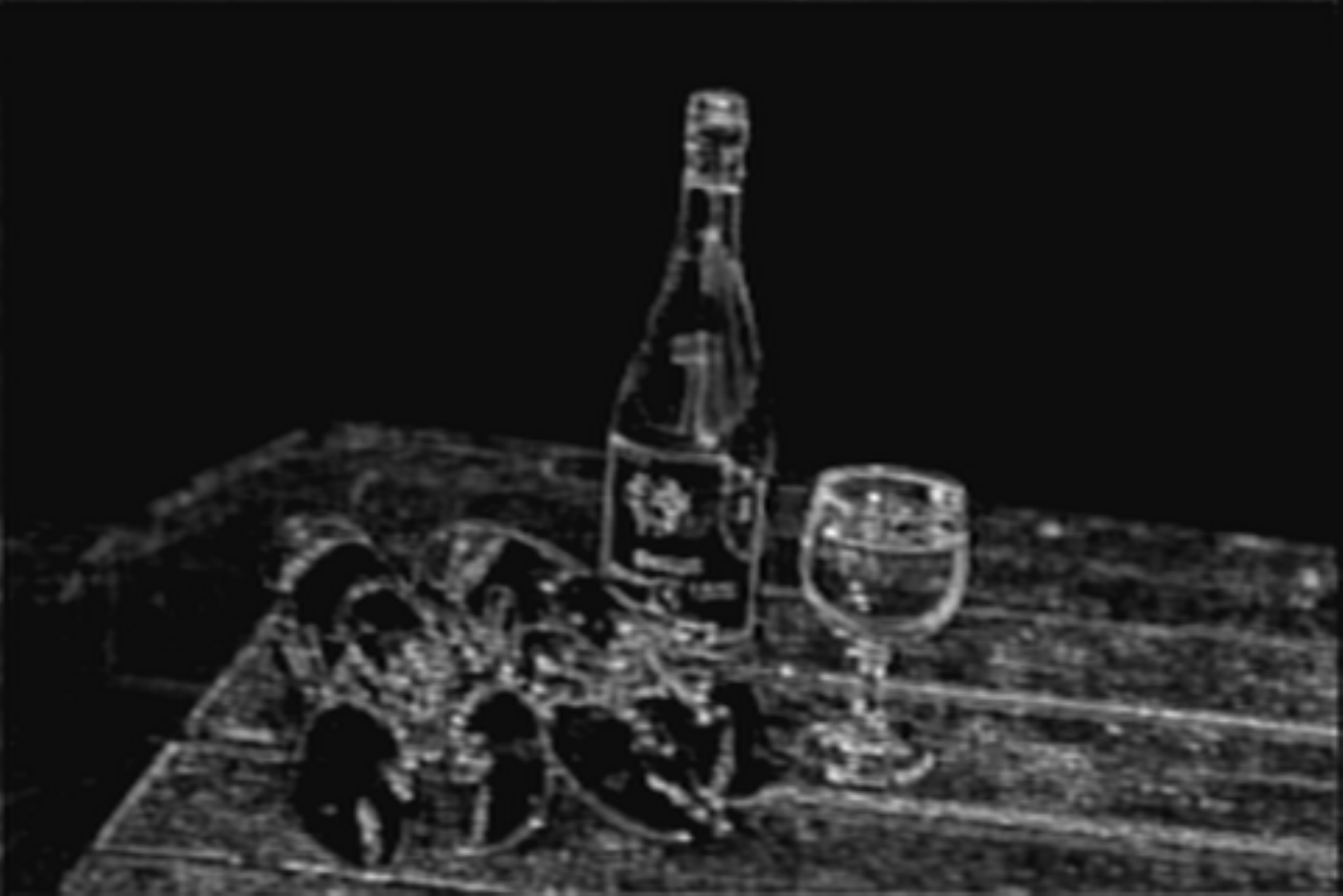}

\captionsetup{labelformat=default}
\setcounter{figure}{1}
    \caption{A visualization of selected convolutional feature maps from VGG network (resized to the input image dimension). Because VGG was optimized for an object classification task, it produces high activation values on objects and their parts.}
    \label{conv_maps}
\end{figure*}


\captionsetup{labelformat=default}

Most of the contour detection methods predict boundaries based purely on color, text, or other low-level features. We can divide these methods into three broad categories: spectral methods, supervised discriminative methods and deep learning based methods.

Spectral methods formulate contour detection problem as an eigenvalue problem. The solution to this problem is then used to reason about the boundaries. The most successful approaches in this genre are the MCG detector~\cite{cArbelaez14}, gPb detector~\cite{Arbelaez:2011:CDH:1963053.1963088}, PMI detector~\cite{crisp_boundaries}, and Normalized Cuts~\cite{Shi97normalizedcuts}. 


Some of the notable discriminative boundary detection methods include sketch tokens (ST)~\cite{LimCVPR13SketchTokens}, structured edges (SE)~\cite{Dollar2015PAMI} and sparse code gradients (SCG)~\cite{ren_nips12}.  While SCG use supervised SVM learning~\cite{Burges98atutorial}, the latter two methods rely on a random forest classifier and models boundary detection as a classification task.

Recently there have been attempts to apply deep learning to the task of boundary detection. SCT~\cite{MYP:ACCV:2014} is a sparse coding approach that reconstructs an image using a learned dictionary and then detect boundaries. Both $N^4$ fields~\cite{DBLP:journals/corr/GaninL14}  and DeepNet~\cite{kivinen2014visual} approaches use Convolutional Neural Networks (CNNs) to predict edges. $N^4$ fields rely on dictionary learning and the use of the Nearest Neighbor algorithm within a CNN framework while DeepNet uses a traditional CNN architecture to predict contours.


The most similar to our approach is DeepEdge~\cite{gberta_2015_CVPR}, which uses a multi-scale bifurcated network to perform contour detection using object-level features. However, we show that our method achieves better results even without the complicated multi-scale and bifurcated architecture of DeepEdge. Additionally, unlike DeepEdge, our system can run in near-real time.


%

In comparison to prior approaches, we offer several contributions. First, we propose to use object-level information to predict boundaries. We argue that such an approach leads to semantic boundaries, which are more consistent with humans reasoning. Second, we avoid feature engineering by learning boundaries from human-annotated data. Finally, we demonstrate excellent results for both low-level and high-level vision tasks. For the boundary detection task, our proposed \HfL boundaries outperform all of the prior methods according to both F-score metrics. Additionally, we show that because \HfL boundaries are based on object-level features, they can be used to improve performance in the high-level vision tasks of semantic boundary labeling, semantic segmentation, and object proposal generation.


\section{Boundary Detection}
\label{boundary_detection}

In this section, we describe our proposed architecture and the specific details on how we predict \HfL boundaries using our method. The detailed illustration of our architecture is presented in Figure~\ref{fig:arch}.

\subsection{Selection of Candidate Contour Points} 

We first extract a set of candidate contour points with a high recall. Due to its efficiency and high recall performance, we use the SE edge detector~\cite{Dollar2015PAMI}. In practice, we could eliminate this step and simply try to predict boundaries at every pixel. However, selecting a set of initial candidate contour points, greatly reduces the computational cost. Since our goal is to build a boundary detector that is both accurate and efficient, we use these candidate points to speed up the computation of our method. 


\subsection{Object-Level Features}


After selecting candidate contour points, we up-sample the original input image to a larger dimension (for example $1100 \times 1100$). The up-sampling is done to minimize the loss of information due to the input shrinkage caused by pooling at the different layers. Afterwards, we feed the up-sampled image through $16$ convolutional layers of the VGG net.

We use the VGG net as our model because it has been trained to recognize a large number of object classes (the $1000$ categories of the ImageNet dataset~\cite{ILSVRCarxiv14}) and thus encodes object-level features that apply to many classes. To preserve specific location information we utilize only the $16$ convolutional layers of the VGG net. We don't use fully connected layers because they don't preserve spatial information, which is crucial for accurate boundary detection. 

We visualize some of the selected convolutional maps in Figure~\ref{conv_maps}. Note the high activation values around the various objects in the images, which confirms our hypothesis that the VGG net encodes object specific information in its convolutional feature maps.

\subsection{Feature Interpolation}

Similarly to~\cite{DBLP:journals/corr/SermanetEZMFL13, DBLP:journals/corr/HariharanAGM14a,long_shelhamer_fcn}, we perform feature interpolation in deep layers. After the up-sampled image passes through all $16$ convolutional layers, for each selected candidate contour point we find its corresponding point in the feature maps. Due to the dimension differences in convolutional maps these correspondences are not exact. Thus we perform feature interpolation by finding the four nearest points and averaging their activation values. This is done in each of the $5504$ feature maps. Thus, this results in a $5504$-dimensional vector for each candidate point. 

We note that the interpolation of convolutional feature maps is the crucial component that enables our system to predict the boundaries efficiently. Without feature interpolation, our method would need to independently process the candidate edge points by analyzing a small image patch around each point, as for example done in DeepEdge~\cite{gberta_2015_CVPR} which feeds one patch at a time through a deep network. However, when the number of candidate points is large (e.g., DeepEdge considers about 15K points at each of 4 different scales), their patches overlap significantly and thus a large amount of computation is wasted by recalculating filter response values over the same pixels. Instead, we can compute the features for all candidate points with a single pass through the network by performing deep convolution over the {\em entire} image (i.e., feeding the entire image rather than one patch at a time) and then by interpolating the convolutional feature maps at the location of each candidate edge point so as to produce its feature descriptor. Thanks to this speedup, our method has a runtime of  $1.2$ seconds (using a K40 GPU), which is better than the runtimes of prior deep-learning based edge detection methods~\cite{Shen_2015_CVPR, DBLP:journals/corr/GaninL14,kivinen2014visual,gberta_2015_CVPR}.

  

\subsection{Learning to Predict Boundaries}

After performing feature interpolation, we feed the $5504$-dimensional feature vectors corresponding to each of the candidate contour points to two fully connected layers that are optimized to the human agreement criterion. To be more precise, we define our prediction objective as a fraction of human annotators agreeing on the presence of the boundary at a particular pixel.  Therefore, a learning objective aims at mimicking the judgement of the human labelers. 

Finally, to detect \HfL boundaries, we accumulate the predictions from the fully connected layers for each of the candidate points and produce a boundary probability map as illustrated in Figure~\ref{fig:arch}.



\subsection{Implementation Details}

In this section, we describe the details behind the training procedure of our model. We use the Caffe library~\cite{jia2014caffe} to implement our network architecture.

In the training stage, we freeze the weights in all of the convolutional layers. To learn the weights in the two fully connected layers we train our model to optimize the least squares error of the regression criterion that we described in the previous subsection. To enforce regularization we set a dropout rate of $0.5$ in the fully connected layers.


Our training dataset includes $80K$ points from the BSDS500 dataset~\cite{MartinFTM01}. As described in the previous subsection, the labels represent the fraction of human annotators agreeing on the boundary presence. We divide the label space into four quartiles, and select an equal number of samples for each quartile to balance the training dataset. In addition to the training dataset, we also sample a hold-out dataset of size $40,000$. We use this for the hard-positive mining~\cite{malisiewicz-iccv11} in order to reduce the number of false-negative predictions.

For the first $25$ epochs we train the network on the original $80,000$ training samples. After the first $25$ epochs, we test the network on the hold-out dataset and detect false negative predictions made by our network. We then augment the original $80,000$ training samples with the false negatives and the same number of randomly selected true negatives. For the remaining $25$ epochs, we train the network on this augmented dataset. 


\subsection{Boundary Detection Results}


In this section, we present our results on the BSDS500 dataset~\cite{MartinFTM01}, which is the most established benchmark for boundary detection. The quality of the predicted boundaries is evaluated using three standard measures: fixed contour threshold (ODS), per-image best threshold (OIS), and average precision (AP).

We compare our approach to the state-of-the-art based on two different sets of BSDS500 ground truth boundaries. First, we evaluate the accuracy by matching each of the predicted boundary pixels with the ground truth boundaries that were annotated by {\em any} of the human annotators. This set of ground truth boundaries is referred to as ``any''.  We present the results for ``any'' ground truth boundaries in the lower half of Table~\ref{any_bsds}. As indicated by the results, \HfL boundaries outperform all the prior methods according to both F-score measures. 



%
%
%

  \begin{table}
    \begin{center}
    \begin{tabular}{ | c  | c  c  c | c |}
    \hline
    {\em Consensus GT} & {\em ODS} & {\em OIS} & {\em AP} & {\em FPS} \\ \hline \hline
	SCG ~\cite{ren_nips12} & 0.6 & 0.64 & 0.56 & 1/280\\ \hline
	DeepNet~\cite{kivinen2014visual}  & 0.61 & 0.64 & 0.61 & $1/5^{\ddagger}$\\ \hline
	PMI~\cite{crisp_boundaries}  & 0.61 & \bf 0.68 & 0.56 & 1/900\\ \hline
	DeepEdge~\cite{gberta_2015_CVPR}  & 0.62 & 0.64 &  0.64& $1/1000^{\ddagger}$\\ \hline
	$N^4$-fields~\cite{DBLP:journals/corr/GaninL14}  & 0.64 & 0.67 & 0.64 & $1/6^{\ddagger}$\\ \hline
	\bf \HfL  & \bf 0.65 & \bf 0.68 & \bf 0.67 & $5/6^{\ddagger}$ \\ \hline  \addlinespace[1ex] \hline
	
        {\em Any GT} & {\em ODS} & {\em OIS} & {\em AP} & {\em FPS} \\ \hline \hline
	SE~\cite{Dollar2015PAMI}  & 0.75 & 0.77 & 0.80 & \bf 2.5\\ \hline
	MCG~\cite{cArbelaez14}  & 0.75 & 0.78 & 0.76 & 1/24\\ \hline
	$N^4$-fields~\cite{DBLP:journals/corr/GaninL14}  & 0.75 & 0.77 & 0.78 & $1/6^{\ddagger}$\\ \hline
	DeepEdge~\cite{gberta_2015_CVPR}  & 0.75 & 0.77 & \bf 0.81 & $1/1000^{\ddagger}$\\ \hline
	MSC~\cite{Sironi_2014_CVPR}  & 0.76 & 0.78 & 0.79 & -\\ \hline
	DeepContour~\cite{Shen_2015_CVPR}  & 0.76 & 0.77 & 0.8 & $1/30^{\ddagger}$\\ \hline
	\bf \HfL  & \bf 0.77 & \bf 0.79 & 0.8 & $5/6^{\ddagger}$\\ 
    \hline
    \end{tabular}
    \end{center}\vspace{-.2cm}
    \caption{Boundary detection results on BSDS500 benchmark. Upper half of the table illustrates the results for ``consensus'' ground-truth criterion while the lower half of the table depicts the results for ``any'' ground-truth criterion. In both cases, our method outperforms all prior methods according to both ODS (optimal dataset scale) and OIS (optimal image scale) metrics. We also report the run-time of our method ($\ddagger$ GPU time) in the FPS column (frames per second), which shows that our algorithm is faster than prior approaches based on deep learning~\cite{Shen_2015_CVPR, DBLP:journals/corr/GaninL14,kivinen2014visual,gberta_2015_CVPR}.}
    \label{any_bsds}
   \end{table}


   \begin{figure}
\centering

\myfigurethreecol{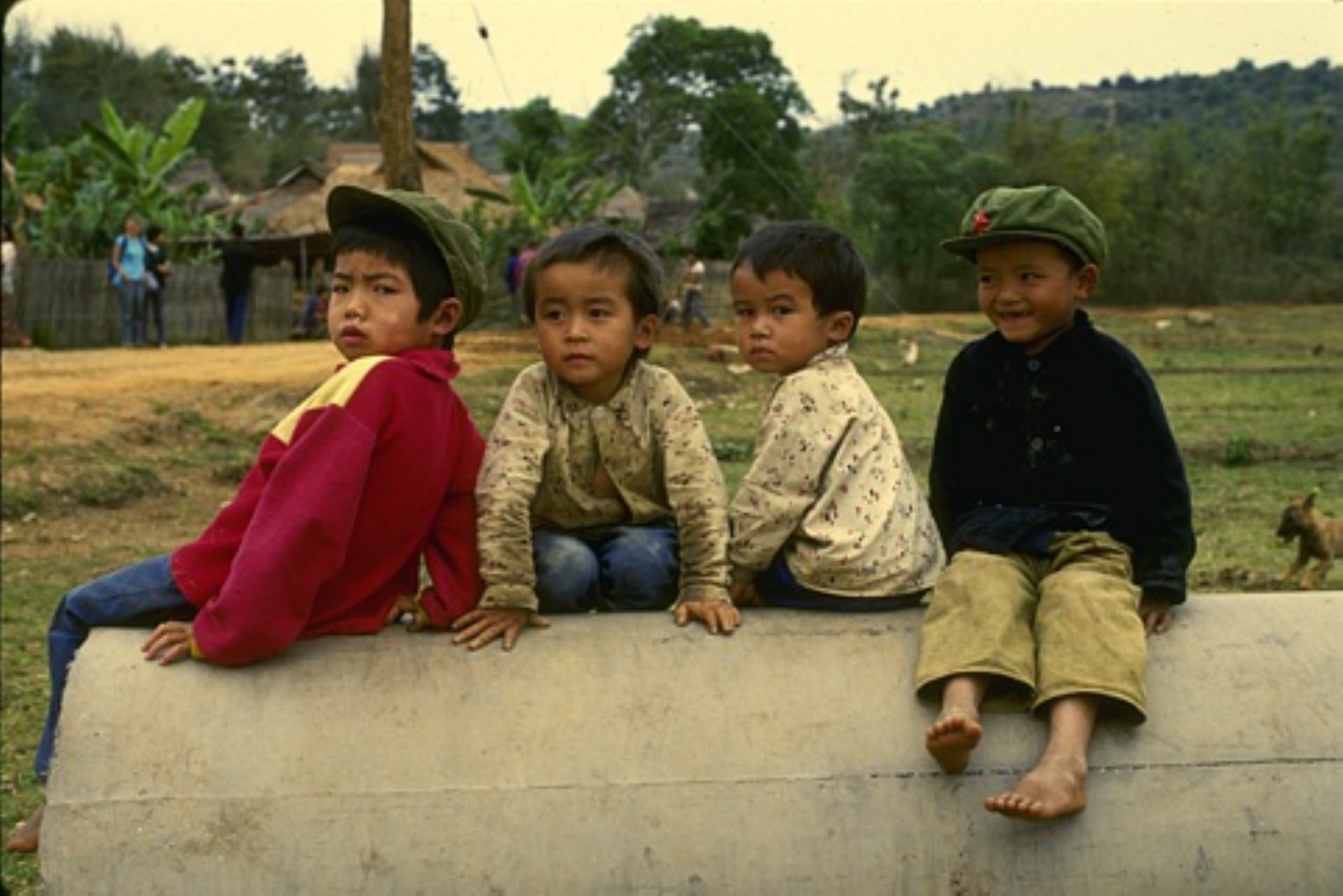}
\myfigurethreecol{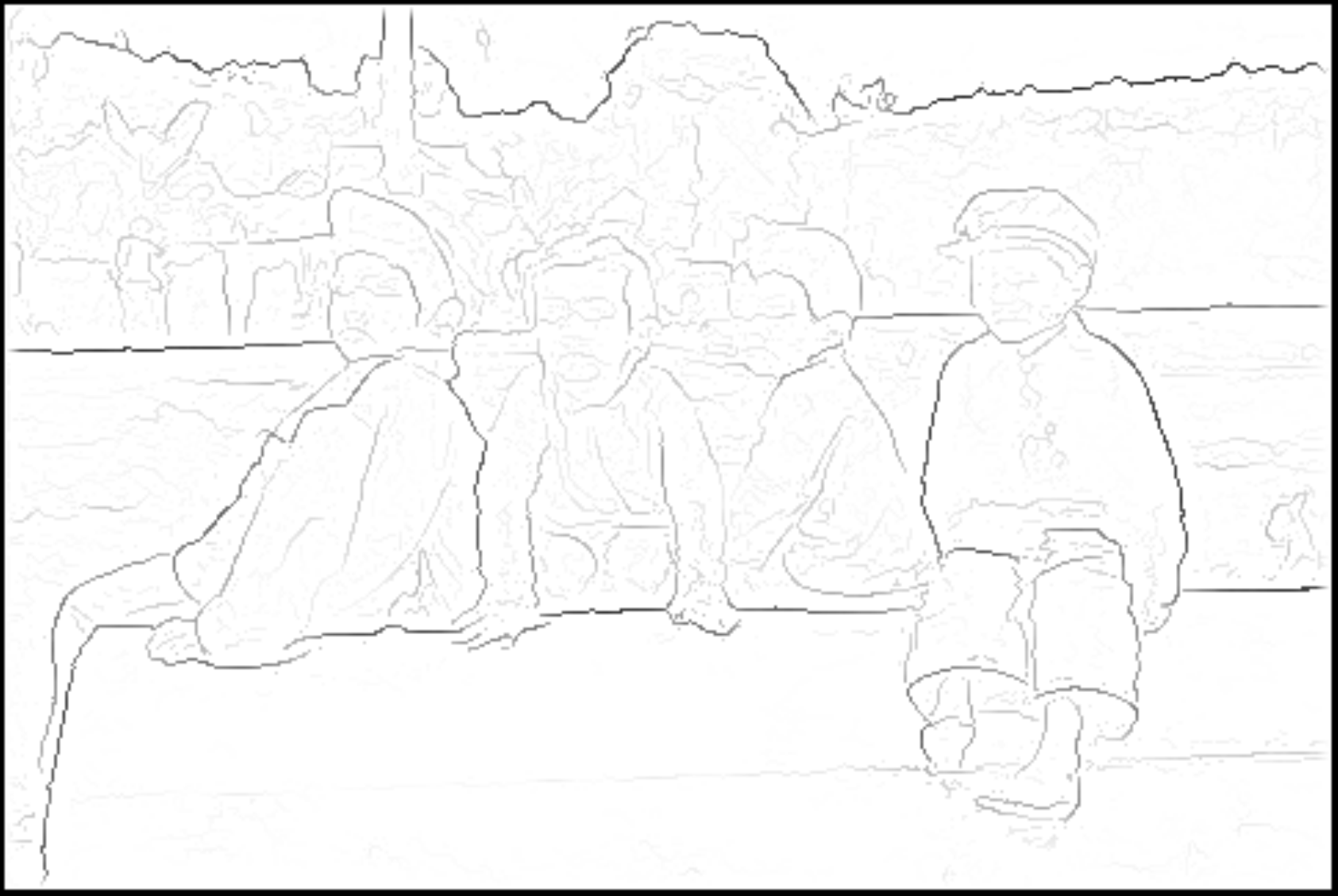}
\myfigurethreecol{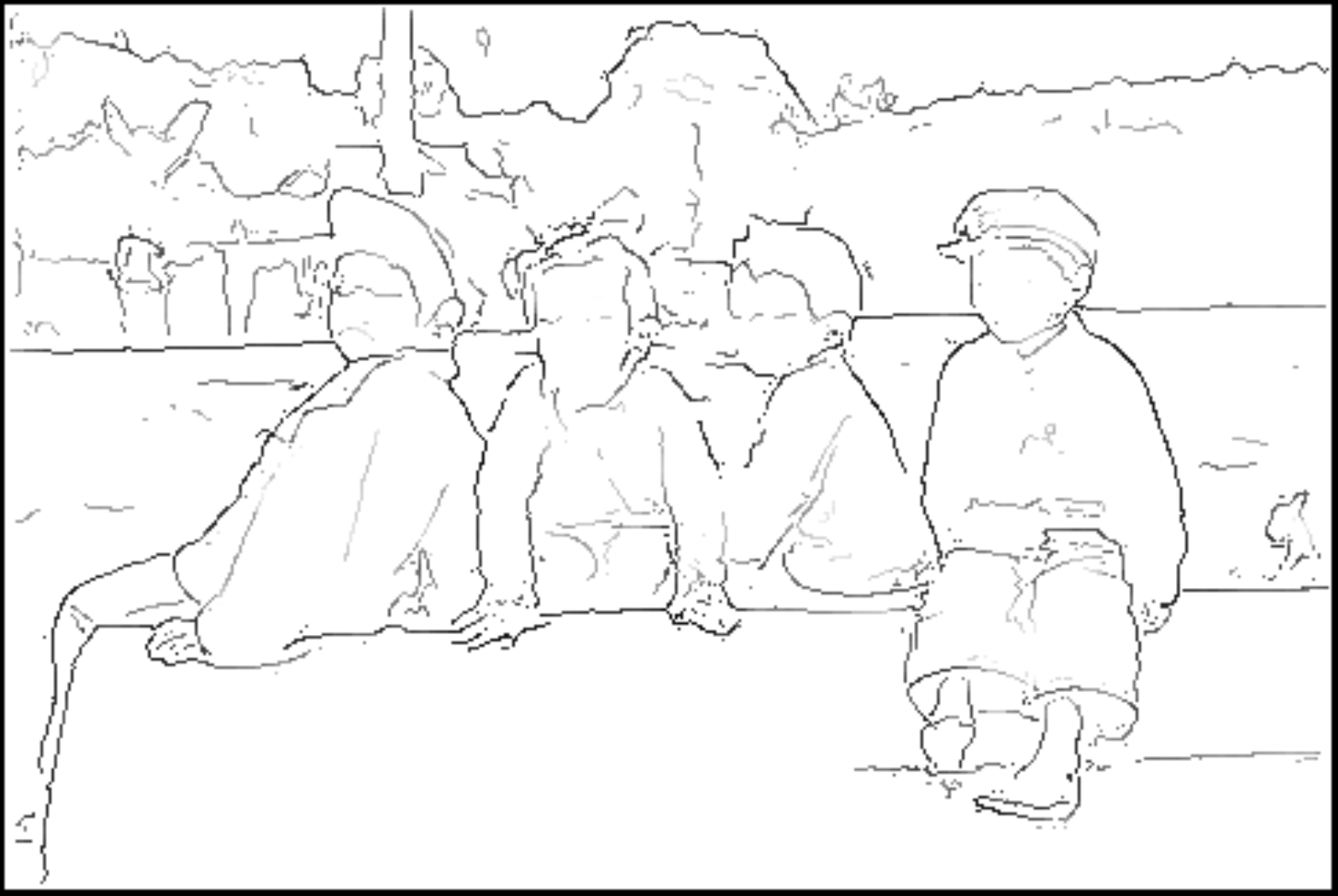}


\myfigurethreecol{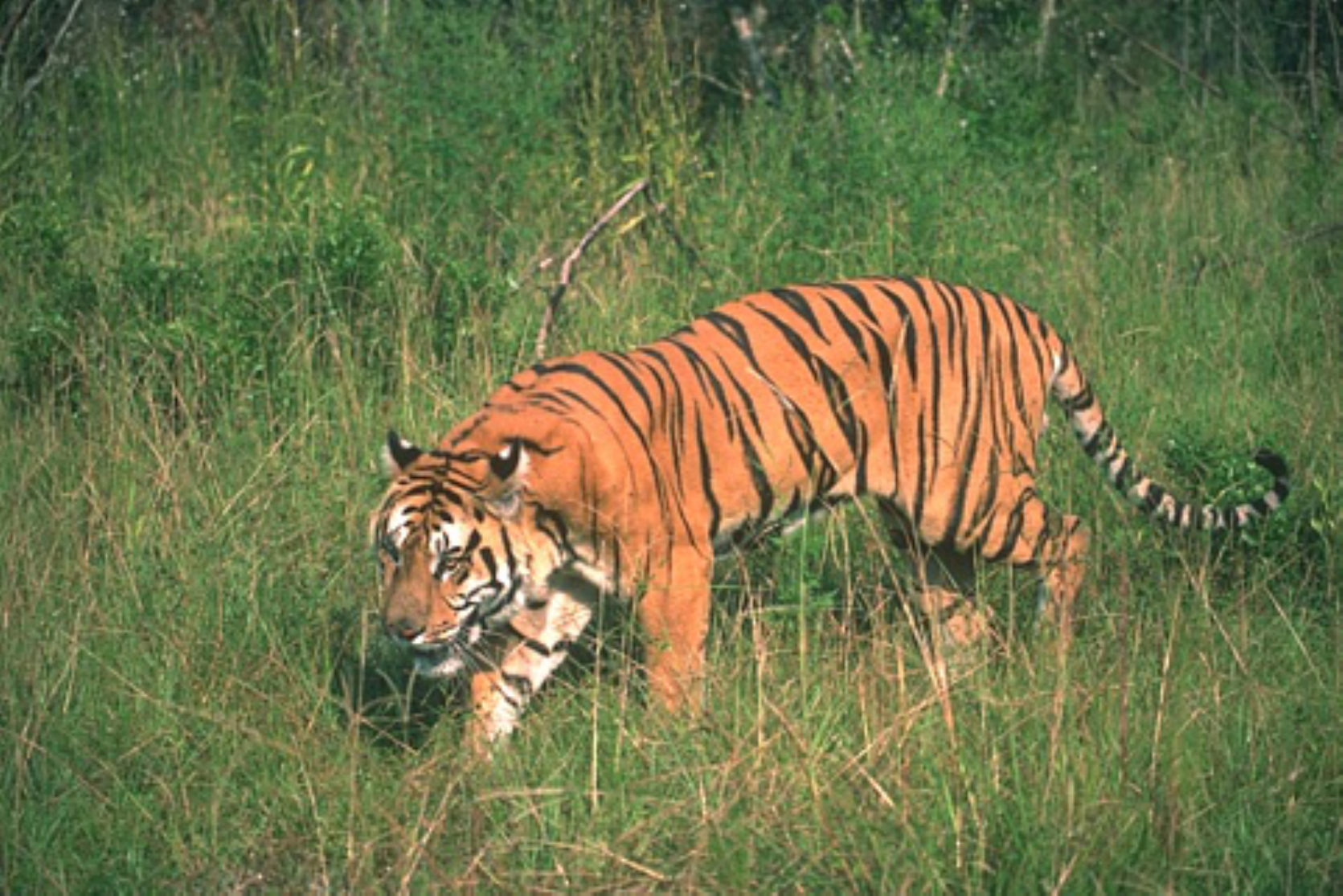}
\myfigurethreecol{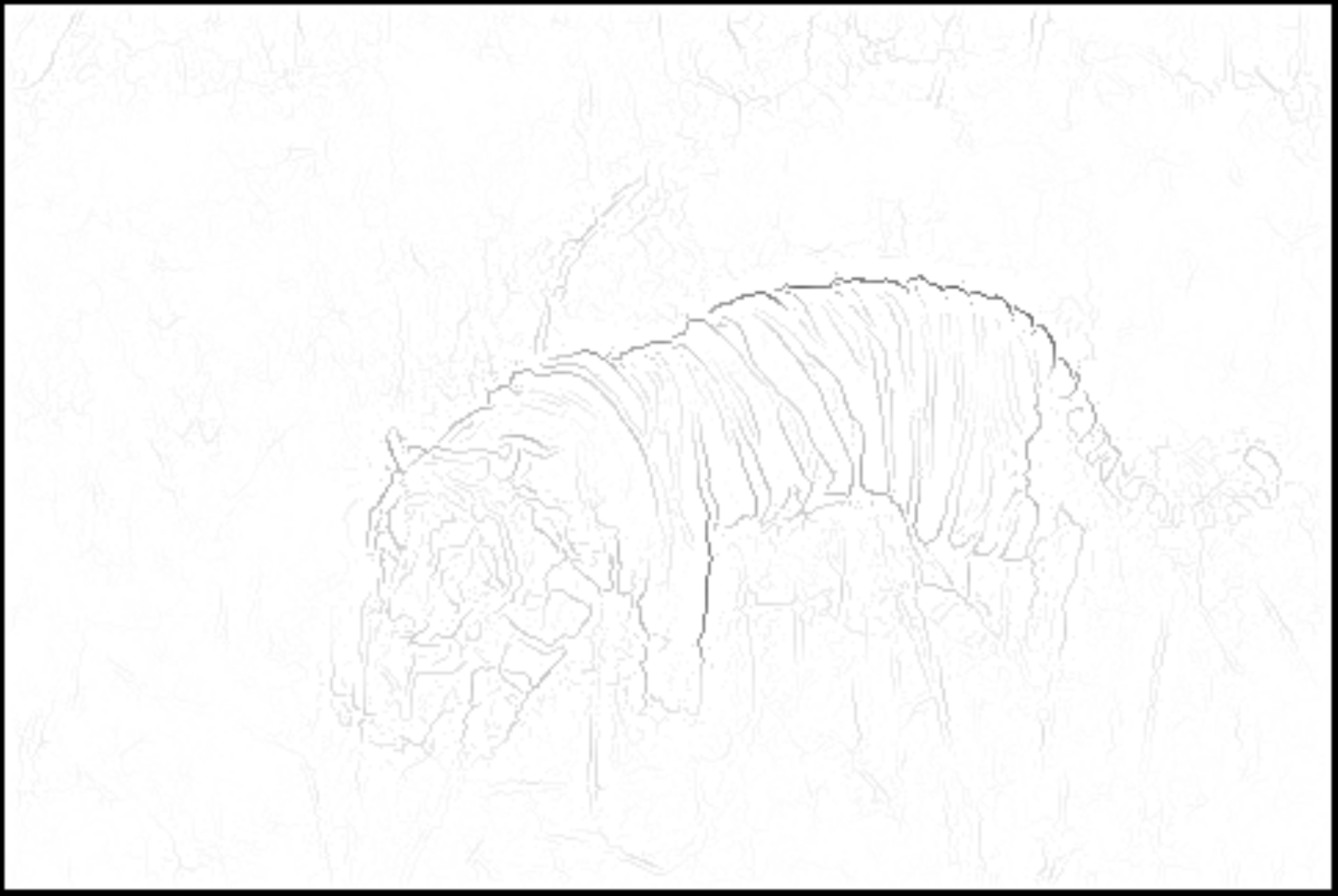}
\myfigurethreecol{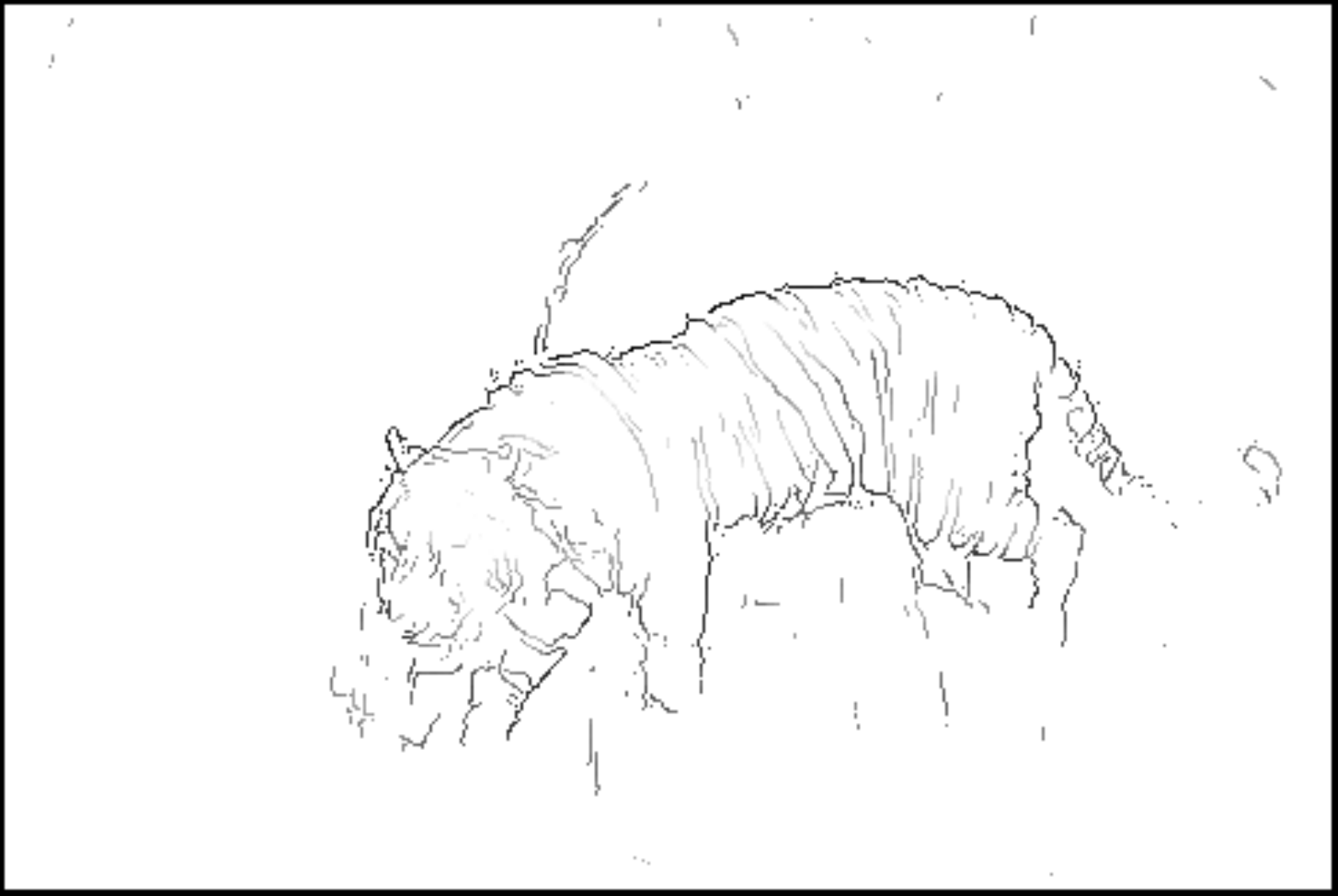}

\captionsetup{labelformat=default}
\setcounter{figure}{4}
    \caption{Qualitative results on the BSDS benchmark. The first column of images represent input images. The second column illustrates SE~\cite{Dollar2015PAMI}, while the third column depicts \HfL boundaries. Notice that SE boundaries are predicted with low confidence if there is no significant change in color between the object and the background. Instead, because our model is defined in terms of object-level features, it can predict object boundaries with high confidence even if there is no significant color variation in the scene.\vspace{-0.2cm}}
    \label{qual_bsds}
\end{figure}

\begin{figure}
\begin{center}
   \includegraphics[width=1\linewidth]{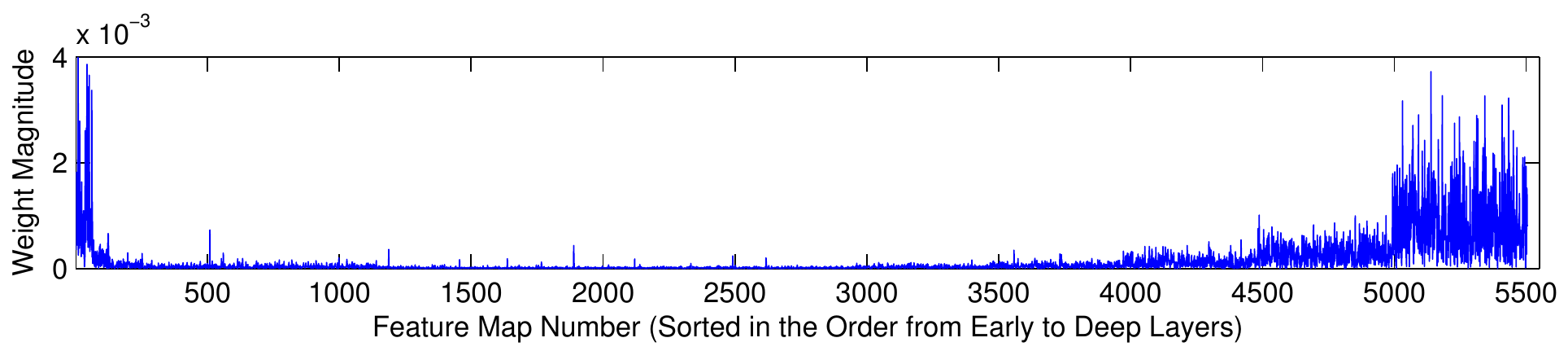}
\end{center}
   \caption{We train a linear regression model and visualize its weight magnitudes in order to understand which features are used most heavily in the boundary prediction (this linear regression is used only for the visualization purposes and not for the accuracy analysis). Note how heavily weighted features lie in the deepest layers of the network, i.e., the layers that are most closely associated with object information.\vspace{-0.4cm}}
\label{fig:feats}
\end{figure}

Recently, there has been some criticism raised about the procedure for boundary detection evaluation on the BSDS500 dataset. One issue with the BSDS500 dataset involves the so called ``orphan'' boundaries: the boundaries that are marked by only one or two human annotators. These ``orphan'' boundaries comprise around $30\%$ of BSDS500 dataset but most of them are considered uninformative. However, the standard evaluation benchmark rewards the methods that predict these boundaries. To resolve this issue we also evaluate our \HfL boundaries on the so called ``consensus'' set of ground truth boundaries. These ``consensus'' boundaries involve only boundaries that are marked by {\em all} of the human annotators and hence, are considered perceptually meaningful. In the upper half of Table~\ref{any_bsds}, we present the results achieved by our method on the ``consensus'' set of the ground truth boundaries. Our \HfL boundaries outperform or tie all the prior methods in each of the three evaluation metrics, thus suggesting that \HfL boundaries are similar to the boundaries that humans annotated. We also report the runtimes in Table~\ref{any_bsds} and note that our method runs faster than previous deep-learning based edge detection systems~\cite{Shen_2015_CVPR, DBLP:journals/corr/GaninL14,kivinen2014visual,gberta_2015_CVPR}.

Our proposed model computes a highly nonlinear function of the 5504-dimensional feature vector of each candidate point. Thus, it is difficult to assess which features are used most heavily by our edge predictor. However, we can gain a better insight by replacing the nonlinear function with a simple linear model. In Fig.~\ref{fig:feats} we show the weight magnitudes of a simple linear regression model (we stress that this linear model is used only for feature visualization purposes). From this Figure, we observe that many important features are located in the deepest layers of the VGG network. As shown in~\cite{DBLP:journals/corr/DonahueJVHZTD13}, these layers encode high-level object information, which confirms our hypothesis that high-level information is useful for  boundary detection.


Finally, we present some qualitative results achieved by our method in Figure~\ref{qual_bsds}. These examples illustrate the effective advantage that \HfL boundaries provide over another state-of-the-art edge detection system, the SE system~\cite{Dollar2015PAMI}. Specifically, observe the parts of the image where there is a boundary that separates an object from the background but where the color change is pretty small. Notice that because the SE boundary detection is based on low-level color and texture features, it captures these boundaries with very low confidence. In comparison, because \HfL boundaries rely on object-level features, it detects these boundaries with high confidence.

\section{High-Level Vision Applications}

In this section, we describe our proposed \textit{Low-for-High} pipeline: using low-level boundaries to aid a number of high-level vision tasks. We focus on the tasks of semantic boundary labeling, semantic segmentation and object proposal generation. We show that using \HfL boundaries improves the performance of state-of-the-art methods in each of these high-level vision tasks.

\subsection{Semantic Boundary Labeling}
\label{sbl}

\setlength{\tabcolsep}{2pt}

     \begin{table*}[t]
     \footnotesize
    \begin{center}
    \begin{tabular}{ | c | c | c | c | c | c | c | c | c | c | c | c | c | c | c | c | c | c | c | c | c ? c |}
    \hline
    Method (Metric) & aero & bike & bird & boat & bottle & bus & car & cat & chair & cow & table & dog & horse & mbike & person & plant & sheep & sofa & train & tv & mean\\ \hline\hline
	InvDet (MF) & 42.6 & 49.5 & 15.7 & 16.8 & 36.7 & 43.0 & 40.8 & 22.6 & 18.1 & 26.6 & 10.2 & 18.0 & 35.2 & 29.4 & 48.2 & 14.3 & 26.8 & 11.2 & 22.2 & 32.0 & 28.0\\ \hline	
	\bf \HfL-FC8 (MF) & 71.6 & 59.6 &  68.0 & 54.1 & 57.2 & 68.0 & 58.8 & 69.3 & 43.3 & 65.8 & 33.3 & 67.9 & 67.5 & 62.2 & 69.0 & 43.8 & 68.5 & 33.9 & 57.7 & 54.8 & 58.7\\ \hline	
	\bf \HfL-CRF (MF) & \bf 73.9 & \bf 61.4 & \bf 74.6 & \bf 57.2 & \bf 58.8 & \bf 70.4 & \bf 61.6 & \bf 71.9 & \bf 46.5 & \bf 72.3 & \bf 36.2 & \bf 71.1 & \bf 73.0 & \bf 68.1 & \bf 70.3 & \bf 44.4 & \bf 73.2 & \bf 42.6 & \bf 62.4 & \bf 60.1 & \bf 62.5\\ \Xhline{4\arrayrulewidth}
	InvDet (AP) & 38.4 & 29.6 & 9.6 & 9.9 & 24.2 & 33.6 & 31.3 & 17.3 & 10.7 & 16.4 & 3.7 & 12.1 & 28.5 & 20.4 & 45.7 & 7.6 & 16.1 & 5.7 & 14.6 & 22.7 & 19.9\\ \hline	
	\bf \HfL-FC8 (AP) & 66.0 & 50.7 & 58.9 & 40.6 & 47.1 & 62.9 & 51.0 & 59.0 & 25.6 & 54.6 & 15.3 & 57.8 & 57.3 & 55.9 & 62.2 & 27.5 & 55.6 & 18.0 & 50.1 & 40.6 & 47.8 \\ \hline	
	\bf \HfL-CRF (AP) & \bf 71.2 & \bf 55.2 & \bf 69.3 & \bf 45.7 & \bf 48.9 & \bf 71.1 & \bf 56.8 & \bf 65.7 & \bf 29.1 & \bf 65.9 & \bf 17.7 & \bf 64.5 & \bf 68.3 & \bf 64.7 & \bf 65.9 & \bf 29.1 & \bf 66.5 & \bf 25.7 & \bf 60.0 & \bf 49.8 & \bf 54.6\\ \hline
    \end{tabular}
    \end{center}
    \caption{Results of semantic boundary labeling on the SBD benchmark using the Max F-score (MF) and Average Precision (AP) metrics. Our method (\HfL) outperforms Inverse Detectors~\cite{BharathICCV2011} for all $20$ categories according to both metrics. Note that using the CRF output to label the boundaries produces better results than using the outputs from the FC8 layer of FCN.}
    \label{sbl_maxf}
   \end{table*}

The task of semantic boundary labeling requires not only to predict the boundaries but also to associate a specific object class to each of the boundaries. This implies that given our predicted boundaries we also need to label them with object class information. We approach this problem by adopting the ideas from the recent work on Fully Convolutional Networks (FCN)~\cite{long_shelhamer_fcn}. Given an input image, we concurrently feed it to our boundary-predicting network (described in Section~\ref{boundary_detection}), and also through the FCN that was pretrained for $20$ Pascal VOC classes and the background class. While our proposed network produces \HfL boundaries, the FCN model predicts class probabilities for each of the pixels. We can then merge the two output maps as follows. For a given boundary point we consider a $9\times 9$ grid around that point from each of the $21$ FCN object-class probability maps. We calculate the maximum value inside each grid, and then label the boundary at a given pixel with the object-class that corresponds to the maximum probability across these $21$ maps. We apply this procedure for each of the boundary points, in order to associate object-class labels to the boundaries. Note that we consider the grids around the boundary pixel because the output of the FCN has a poor localization, and considering the grids rather than individual pixels leads to higher accuracy.  

We can also merge \HfL boundaries with the state-of-the-art DeepLab-CRF segmentation~\cite{DBLP:journals/corr/ChenPKMY14} to obtain higher accuracy. We do this in a similar fashion as just described. First, around a given boundary point we extract a $9 \times 9$ grid from the DeepLab-CRF segmentation. We then compute the mode value in the grid (excluding the background class), and use the object-class corresponding to the mode value as a label for the given boundary point. We do this for each of the boundary points. By merging \HfL boundaries and the output of FCN or DeepLab-CRF, we get semantic boundaries that are highly localized and also contain object-specific information.

\subsubsection{Semantic Boundary Labeling Results}

In this section, we present semantic boundary labeling results on the SBD dataset~\cite{BharathICCV2011}, which includes ground truth boundaries that are also labeled with one of $20$ Pascal VOC classes. The boundary detection accuracy for each class is evaluated using the maximum F-score (MF), and average precision (AP) measures.

Labeling boundaries with the semantic object information is a novel and still relatively unexplored problem. Therefore, we found only one other approach (Inverse Detectors) that tried to tackle this problem~\cite{BharathICCV2011}. The basic idea behind Inverse Detectors consists of several steps. First, generic boundaries in the image are detected. Then, a number of object proposal boxes are generated. These two sources of information are then used to construct the features. Finally, a separate classifier is used to label the boundaries with the object-specific information.


Table~\ref{sbl_maxf} shows that our approach significantly outperforms Inverse Detectors according to both the maximum F-score and the average precision metrics for all twenty categories. As described in Section~\ref{sbl} we evaluate the two variants of our method. Denoted by \HfL-FC8 is the variant for which we label \HfL boundaries with the outputs from the last layer (FC8) of the pretrained FCN. We denote with \HfL-CRF the result of labeling our boundaries with the output from the DeepLab-CRF~\cite{DBLP:journals/corr/ChenPKMY14}. Among these two variants, we show that the latter one produces better results. This is expected since the CRF framework enforces spatial coherence in the semantic segments. 

In Figure~\ref{qual_sbl}, we present some of the qualitative results produced by our method. We note that even with multiple objects in the image, our method successfully recognizes and localizes boundaries of each of the classes.


\captionsetup{labelformat=empty}

\begin{figure}
\centering

\myfigurethreecol{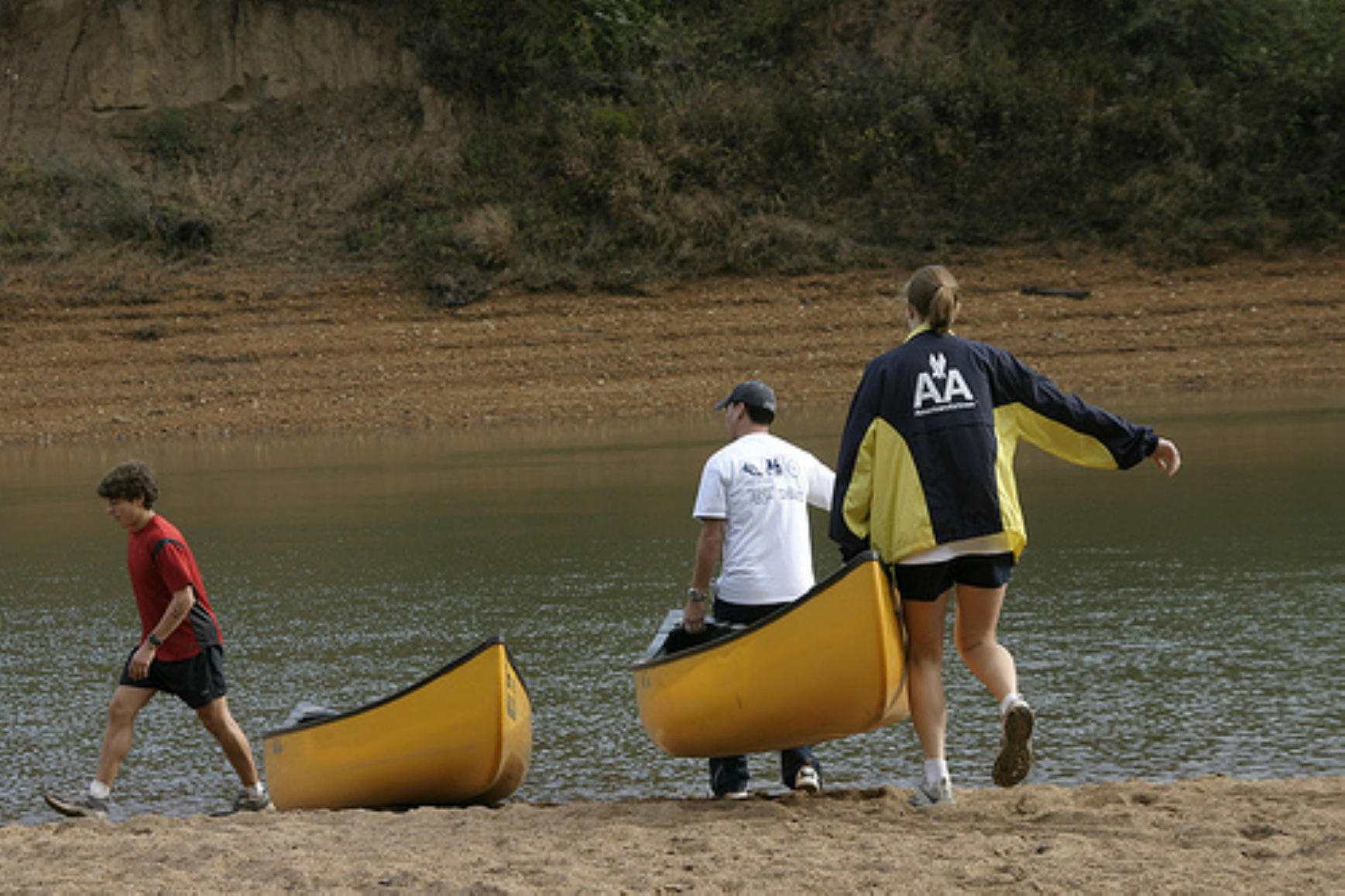}
\myfigurethreecol{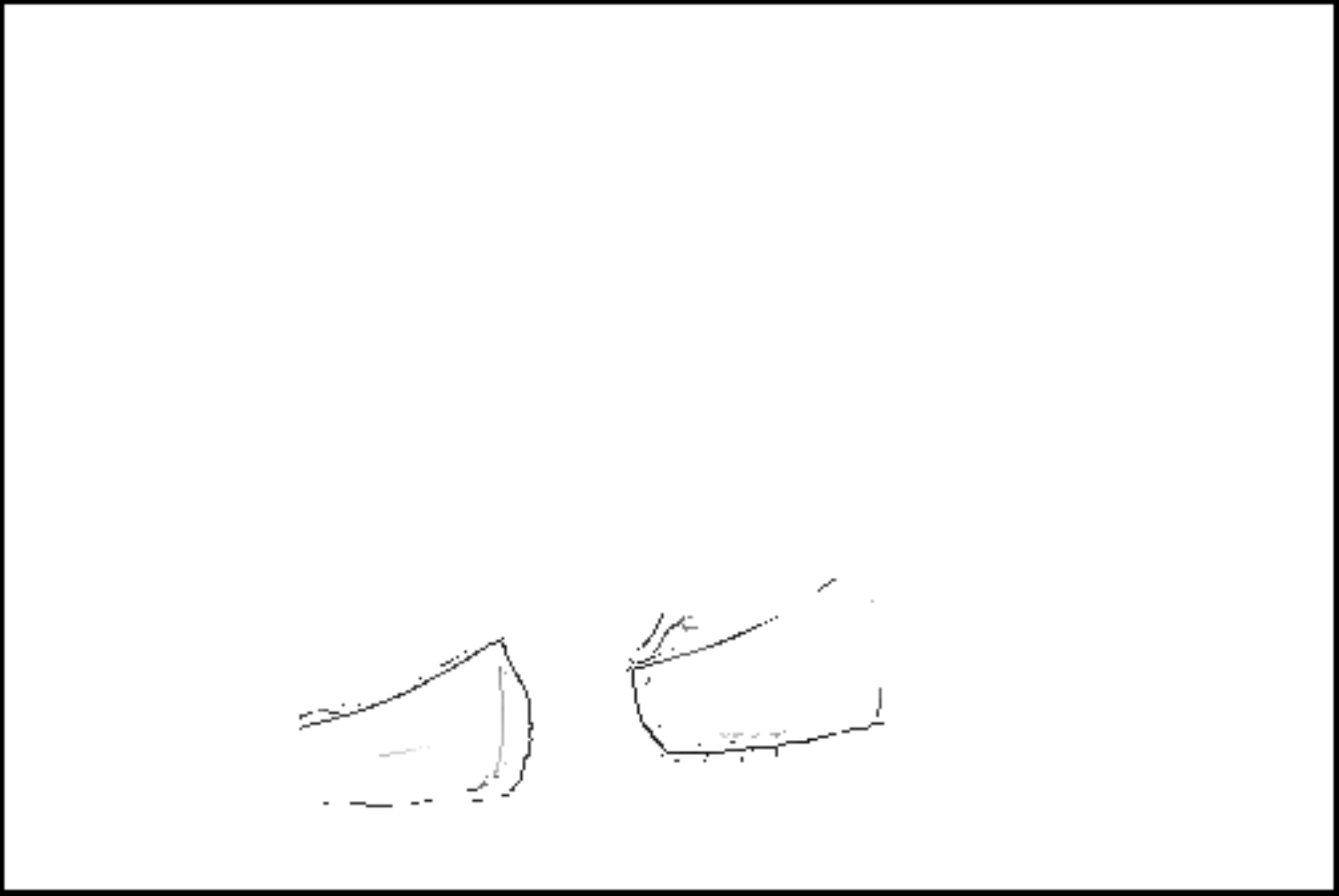}
\myfigurethreecol{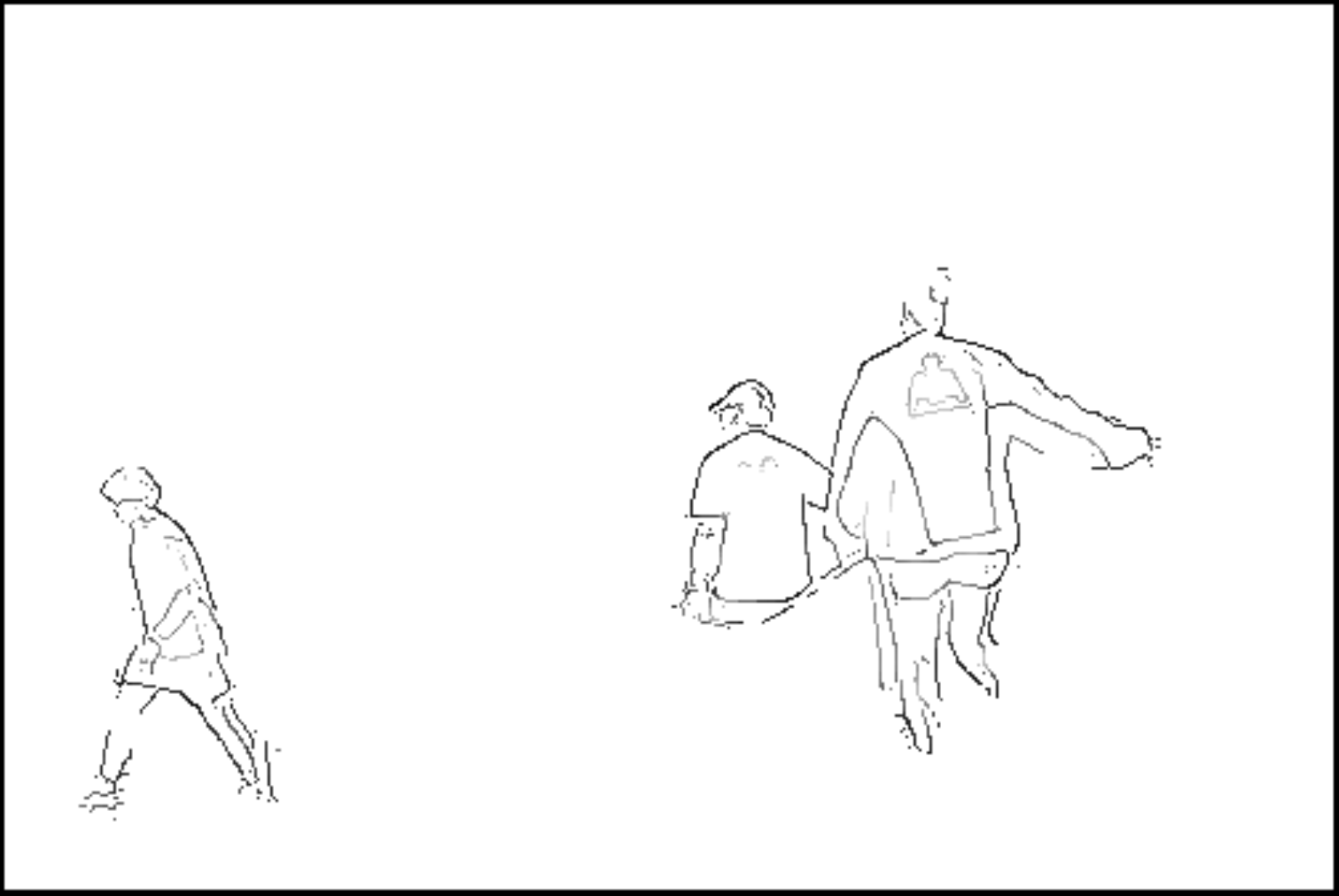}




\myfigurethreecol{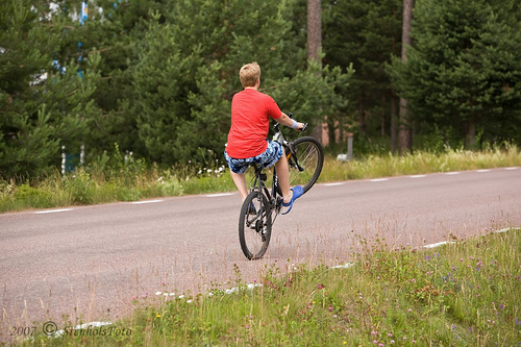}
\myfigurethreecol{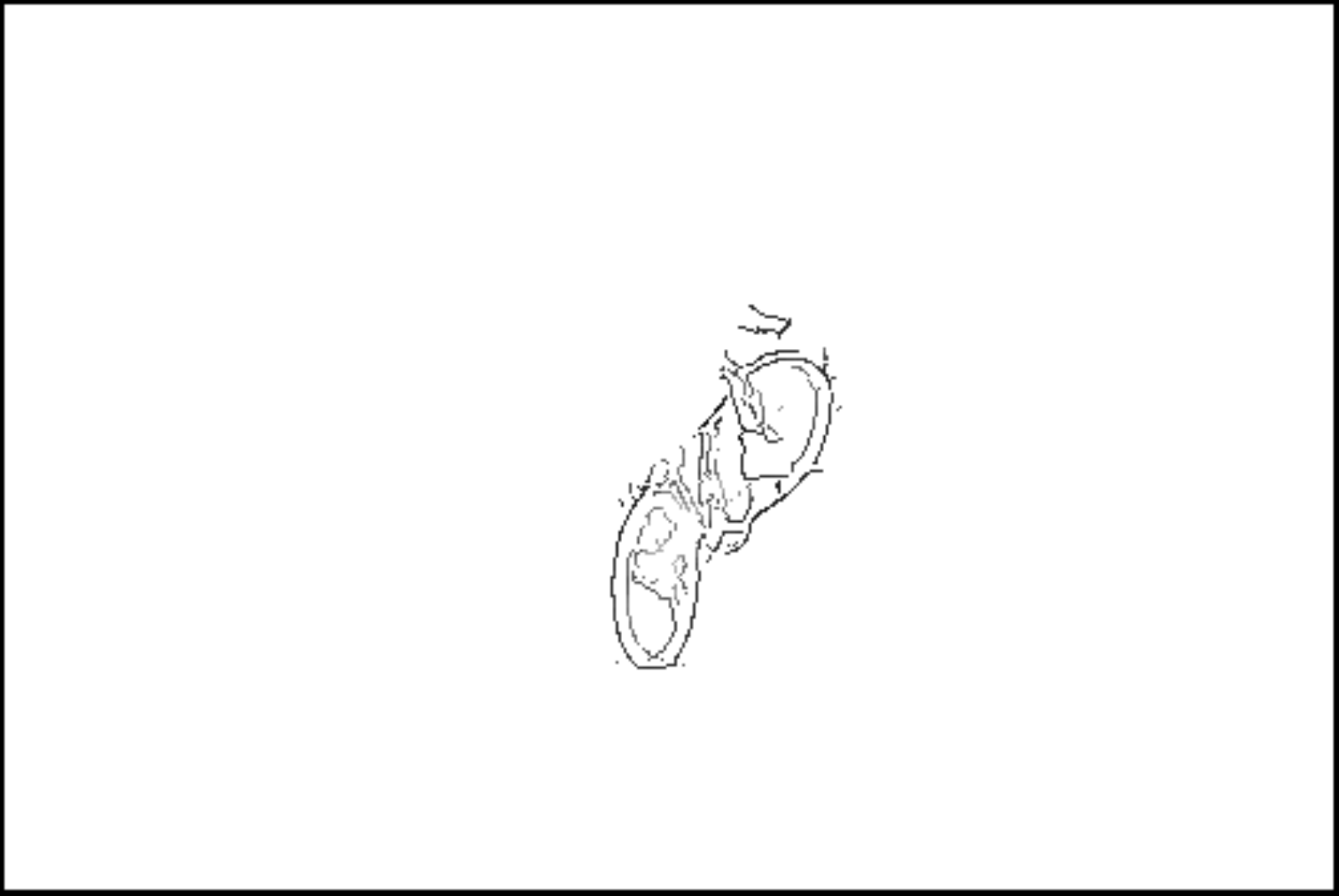}
\myfigurethreecol{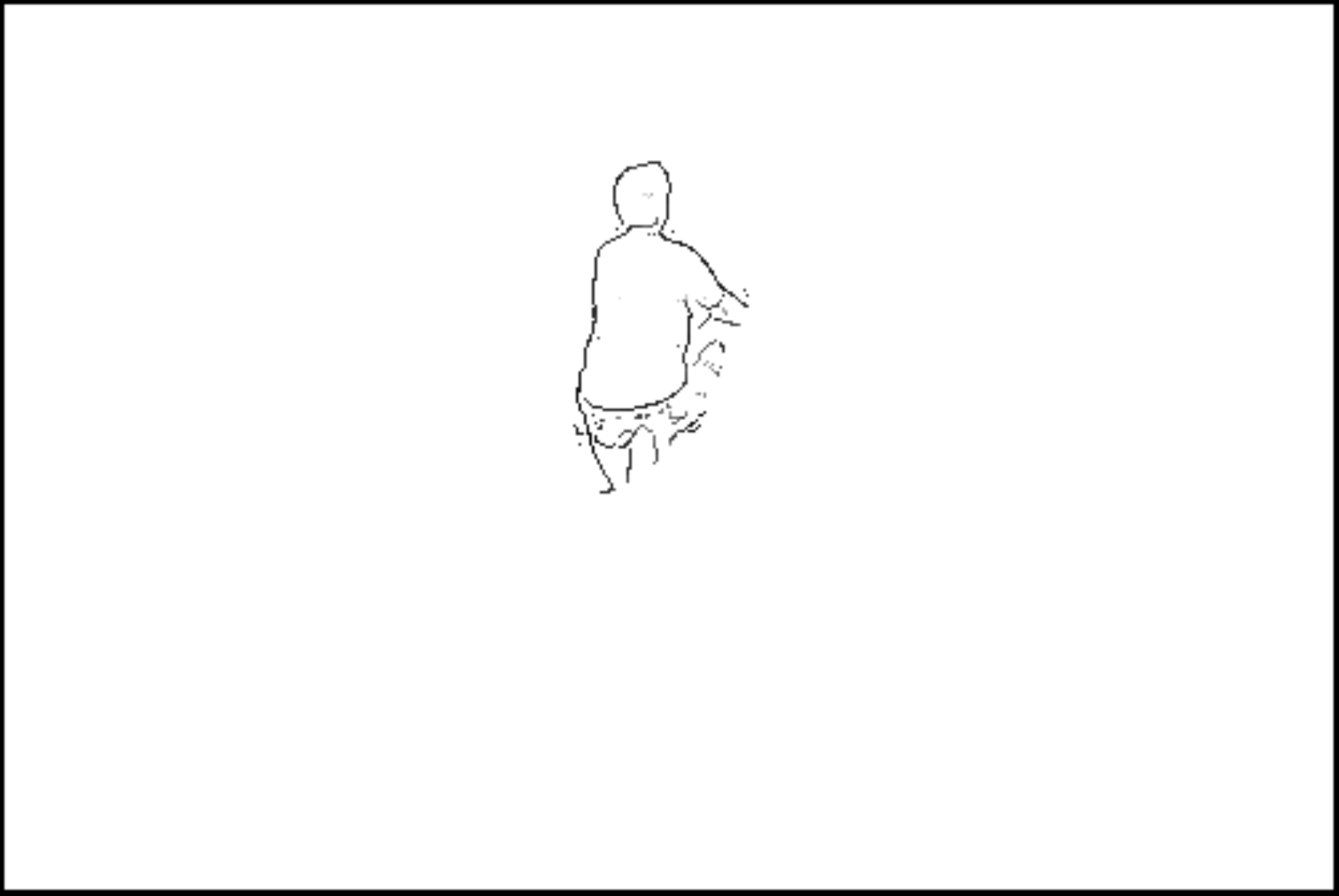}

\captionsetup{labelformat=default}
\setcounter{figure}{6}
    \caption{A visualization of the predicted semantic boundary labels. Images in the first column are input examples. Columns two and three show semantic \HfL boundaries of different object classes. Note that even with multiple objects appearing simultaneously, our method outputs precise semantic boundaries.\vspace{-0.2cm}}
    \label{qual_sbl}
\end{figure}

\subsection{Semantic Segmentation}

For the semantic segmentation task, we propose to enhance the DeepLab-CRF~\cite{DBLP:journals/corr/ChenPKMY14} with our predicted \HfL boundaries. DeepLab-CRF is a system comprised of a Fully Convolutional Network (described in Section~\ref{sbl}) and a dense CRF applied on top of FCN predictions.

Specifically, in the CRF, the authors propose to use a Gaussian kernel and a bilateral term including position and color terms as the CRF features (see~\cite{DBLP:journals/corr/ChenPKMY14}). While in most cases the proposed scheme works well, DeepLab-CRF sometimes produces segmentations that are not spatially coherent, particularly for images containing small object regions. 

We propose to address this issue by adding features based on our predicted \HfL boundaries in the CRF framework. Note that we use predicted boundaries from Section~\ref{boundary_detection} and not the boundaries labeled with the object information that we obtained in Section~\ref{sbl}. We use the Normalized Cut~\cite{Shi97normalizedcuts} framework to generate our features. 

\captionsetup{labelformat=default}

  \setlength{\tabcolsep}{2pt}

     \begin{table*}
     \scriptsize
    \begin{center}
    \begin{tabular}{ | c | c | c | c | c | c | c | c | c | c | c | c | c | c | c | c | c | c | c | c | c | c ? c |}
    \hline
     Metric & Method (Dataset) & aero & bike & bird & boat & bottle & bus & car & cat & chair & cow & table & dog & horse & mbike & person & plant & sheep & sofa & train & tv & mean\\ \hline
        \multirow{4}{*}{PP-IOU}
        	& DL-CRF (VOC)& \bf 78.6 & 41.1 & \bf 83.5 & \bf 75.3 & 72.9 & 83.1 & \bf 76.6 & \bf 80.8 & \bf 37.8 & \bf 72.1 & 66.5 & \bf 64.7 & 65.8 & \bf 75.7 & \bf 80.5 & \bf 34.4 & \bf 75.9 & \bf 47.4 & 86.6 & \bf 77.9 & \bf 68.9\\ 	
	& \bf DL-CRF+\HfL (VOC) & 77.9 & \bf 41.2 & 83.1 & 74.4 & \bf 73.2 & \bf 85.5 & 76.1 & 80.6 & 35.7 & 71.0 & \bf 66.6 & 64.3 & \bf 65.9 & 75.2 & 80.2 & 32.8 & 75.2 & 47.0 & \bf 87.1 & \bf 77.9 & 68.5\\  \cline{2-23}
	& DL-CRF (SBD)& 74.2 & 68.0 & \bf 81.9 & 64.6 & \bf 71.8 & 86.3 & \bf 78.3 & \bf 84.3 & \bf 41.6 & \bf 78.0 & 49.9 & \bf 82.0 & \bf 78.5 & 77.1 & 80.1 & \bf 54.3 & \bf 75.6 & \bf 49.8 & \bf 79.5 & 70.1 & \bf 71.4\\ 	
	& \bf DL-CRF+\HfL (SBD)& \bf 75.1 & \bf 69.2 & 81.6 & \bf 64.8 & 71.3 & \bf 86.4 & 78.1 & 84.1 & 41.2 & 77.8 & \bf 50.4 & 81.6 & 78.2 & \bf 78.5 & \bf 80.7 & 53.8 & 74.9 & 49.1 & \bf 79.5 & \bf 70.4 & \bf 71.4\\  \Xhline{4\arrayrulewidth}
	 \multirow{4}{*}{PI-IOU}
	 & DL-CRF (VOC)& 46.1 & \bf 28.0 & 48.5 & 54.5 & 45.5 & 57.6 & 34.1 & \bf 47.3 & 19.5 & \bf 61.4 & \bf 41.6 & 42.5 & 34.4 & 61.8 & \bf 62.1 & \bf 22.1 & 50.5 & 41.0 & 61.2 & 31.9 & 44.6\\	
	& \bf DL-CRF+\HfL (VOC) & \bf 47.5 & 27.6 & \bf 50.4 & \bf 63.5 & \bf 47.7 & \bf 57.9 & \bf 38.7 & 47.2 & \bf 21.1 & 57.3 & 41.2 & \bf 43.7 & \bf 36.0 & \bf 66.4 & 61.1 & 21.3 & \bf 53.9 & \bf 42.1 & \bf 70.9 & \bf 34.6 & \bf 46.5\\ \cline{2-23}
	& DL-CRF (SBD) & 59.4 & 36.5 & 58.0 & 38.6 & 32.0 & 58.1 & 44.7 & 59.6 & 25.8 & 51.8 & 28.1 & 59.0 & 46.9 & 50.3 & 61.8 & 22.2 & 45.9 & 33.4 & 62.1 & 41.0 & 45.8\\
	& \bf DL-CRF+\HfL (SBD) & \bf 63.4 & \bf 42.5 & \bf 58.4 & \bf 41.3  & \bf 32.5 & \bf 61.2 & \bf 45.7 & \bf 61.4 & \bf 28.4 & \bf 55.5 & \bf 31.5 & \bf 61.4 &  \bf 51.8 & \bf 54.6 & \bf 62.1 & \bf 24.9 & \bf 52.6 & \bf 34.2 & \bf 67.1 & \bf 45.1 & \bf 48.8\\ \hline

      \end{tabular}
    \end{center}
    \caption{Semantic segmentation results on the SBD and VOC 2007 datasets. We measure the results according to PP-IOU (per pixel) and PI-IOU (per image) evaluation metrics. We denote the original DeepLab-CRF system and our proposed modification as DL-CRF and DL-CRF+\HfL, respectively. According to the PP-IOU metric, our proposed features (DL-CRF+\HfL) yield almost equivalent results as the original DeepLab-CRF system. However, based on PI-IOU metric, our proposed features improve the mean accuracy by $3\%$ and $1.9\%$ on SBD and VOC 2007 datasets respectively.}
    \label{pp_iou}
   \end{table*}

First, we construct  a pixel-wise affinity matrix $\bf W$ using our \HfL boundaries. We measure the similarity between two pixels as:
\begin{equation}
W_{ij}=\exp{(-\max_{p \in \overline{ij}}\{\frac{M(p)^2}{\sigma^2}\})} \nonumber
\end{equation} 
 where $W_{ij}$ represents the similarity between pixels $i$ and $j$, $p$ denotes the boundary point along the line segment $\overline{ij}$ connecting pixels $i$ and $j$, $M$ depicts the magnitude of the boundary at pixel $p$, and $\sigma$ denotes the smoothness parameter, which is usually set to $14\%$ of the maximum boundary value in the image. 
 
The intuitive idea is that two pixels are similar (i.e. $W_{ij}=1$) if there is no boundary crossing the line connecting these two pixels (i.e. $M(p)=0\quad \forall p \in \overline{ij}$) or if the boundary strength is low. We note that it is not necessary to build a full affinity matrix $W$. We build a sparse affinity matrix connecting every pair of pixels $i$ and $j$ that have distance $5$ or less from each other.
 
 After building a boundary-based affinity matrix $\bf W$  we set $D_{ii}=\sum_{i \neq j} W_{ij}$ and compute eigenvectors $\bf v$ of the generalized eigenvalue system:
 
 $$(\bf{D}-\bf{W})\bf{v}=\lambda D\bf{v}$$
 
 We then resize the eigenvectors $v$ to the original image dimensions, and use them as additional features to the CRF part of DeepLab-CRF system. In our experiments, we use the $16$ eigenvectors corresponding to the smallest eigenvalues, which results in $16$ extra feature channels.
 
Note that the eigenvectors contain soft segmentation information. Because \HfL boundaries predict object-level contours with high confidence, the eigenvectors often capture regions corresponding to objects. We visualize a few selected eigenvectors in Figure~\ref{eigv}. In the experimental section, we demonstrate that our proposed features make the output produced by DeepLab-CRF more spatially coherent and improve the segmentation accuracy according to one of the metrics.

We also note that our proposed features are applicable to any generic method that incorporates CRF. For instance, even if DeepLab-CRF used an improved DeepLab network architecture, our features would still be beneficial because they contribute directly to the CRF part and not the DeepLab network part of the system.

\subsubsection{Semantic Segmentation Results}

\captionsetup{labelformat=empty}

\begin{figure}
\centering


\myfigurethreecol{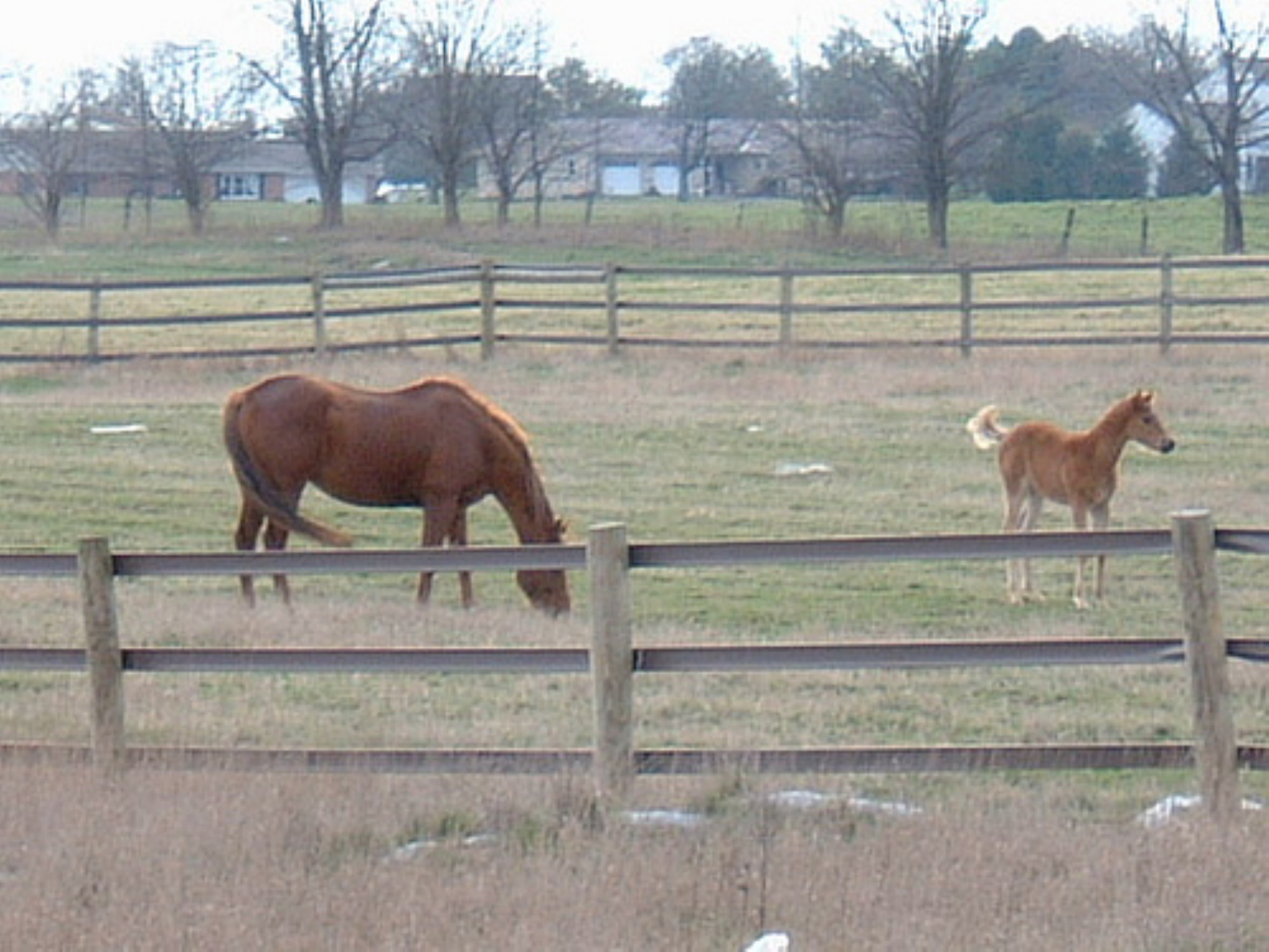}
\myfigurethreecol{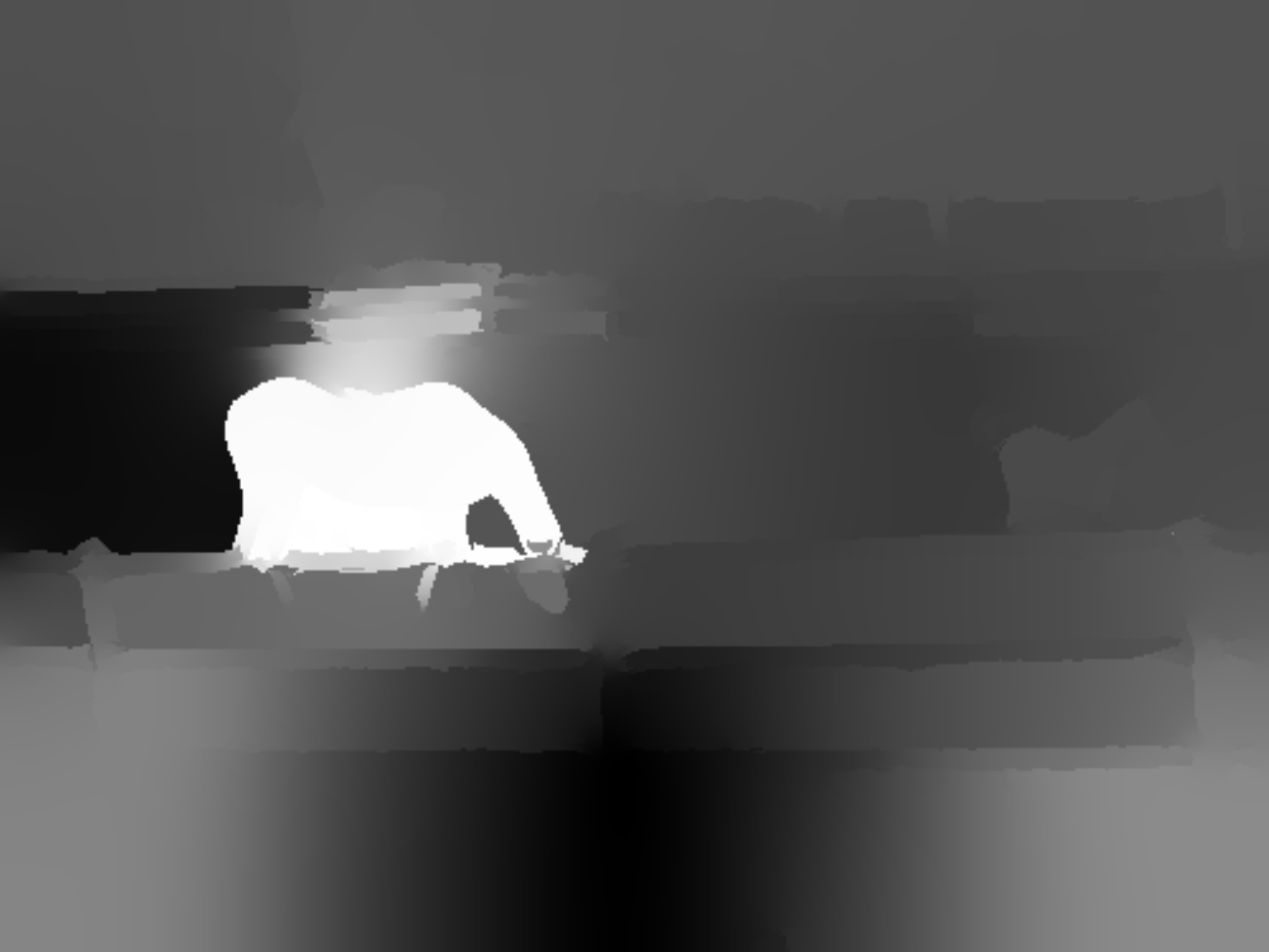}
\myfigurethreecol{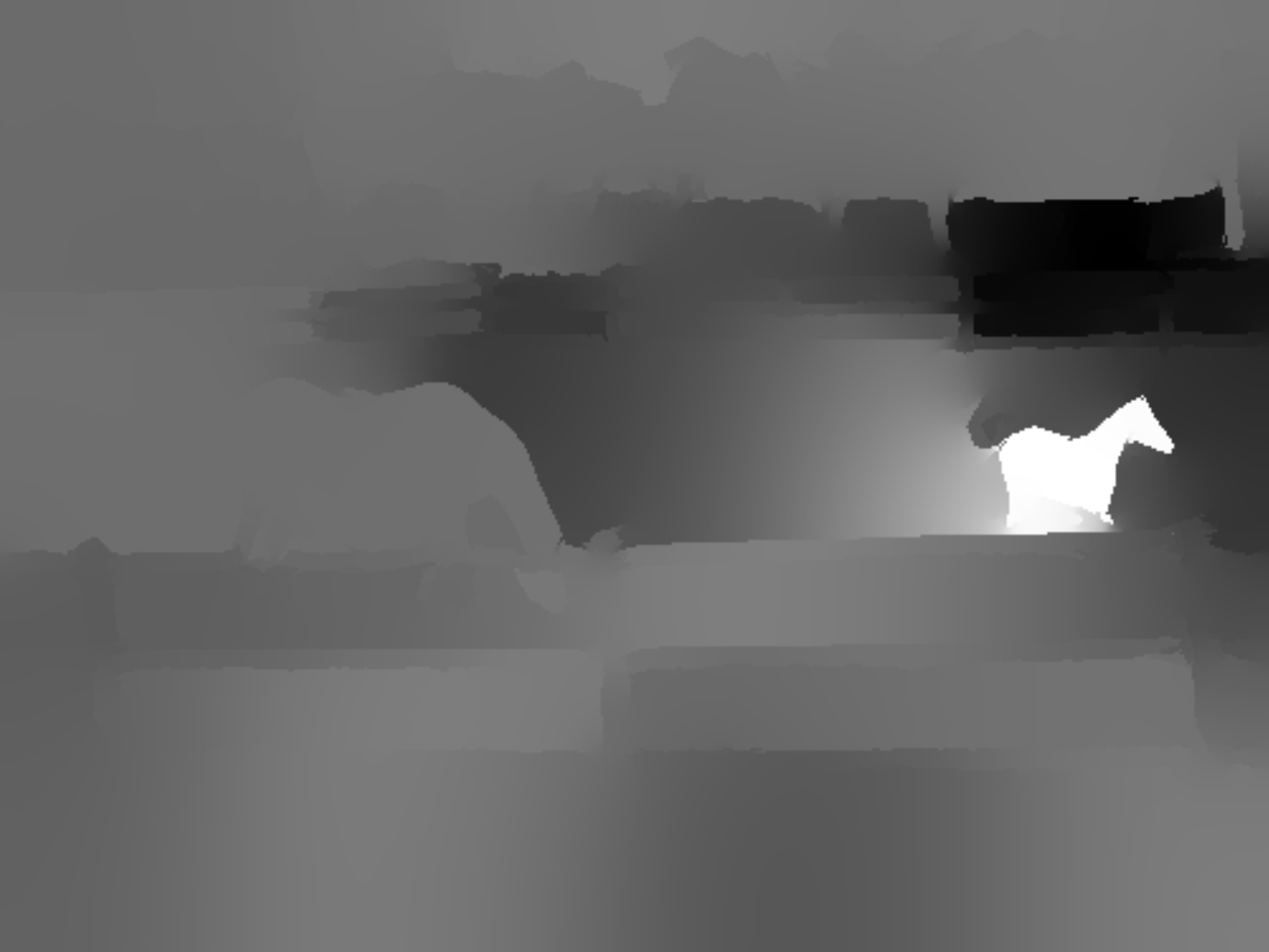}

\myfigurethreecol{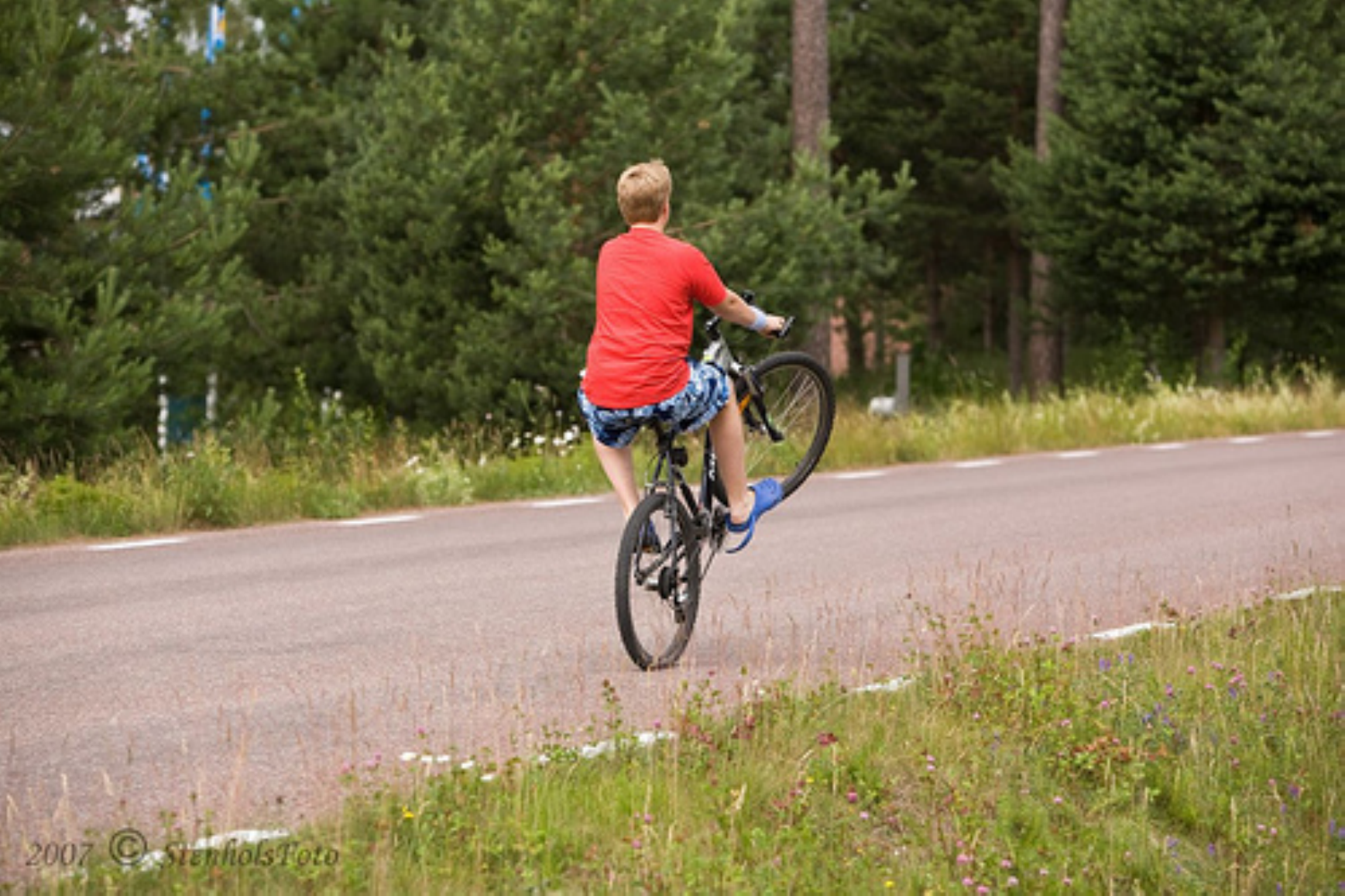}
\myfigurethreecol{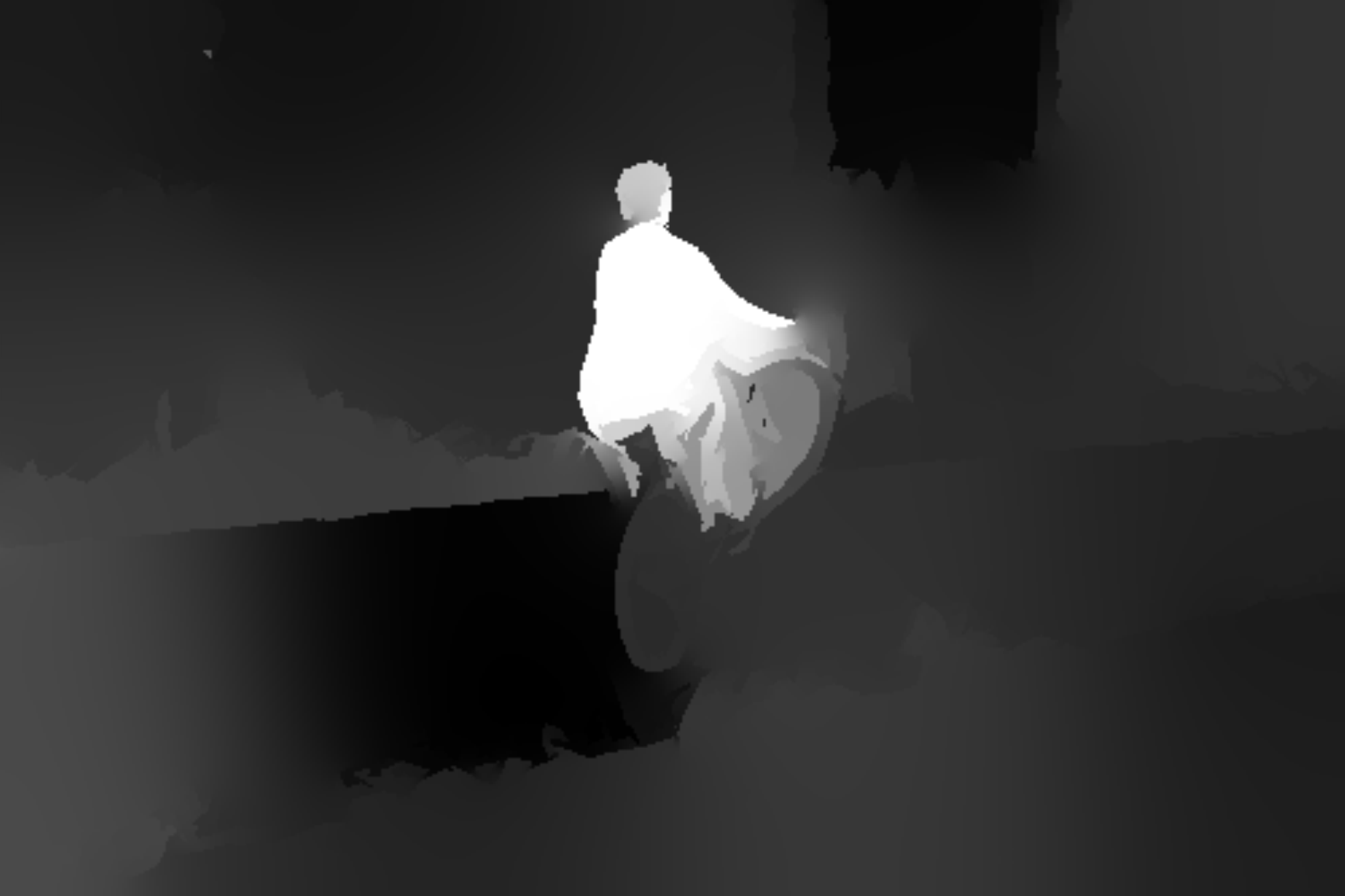}
\myfigurethreecol{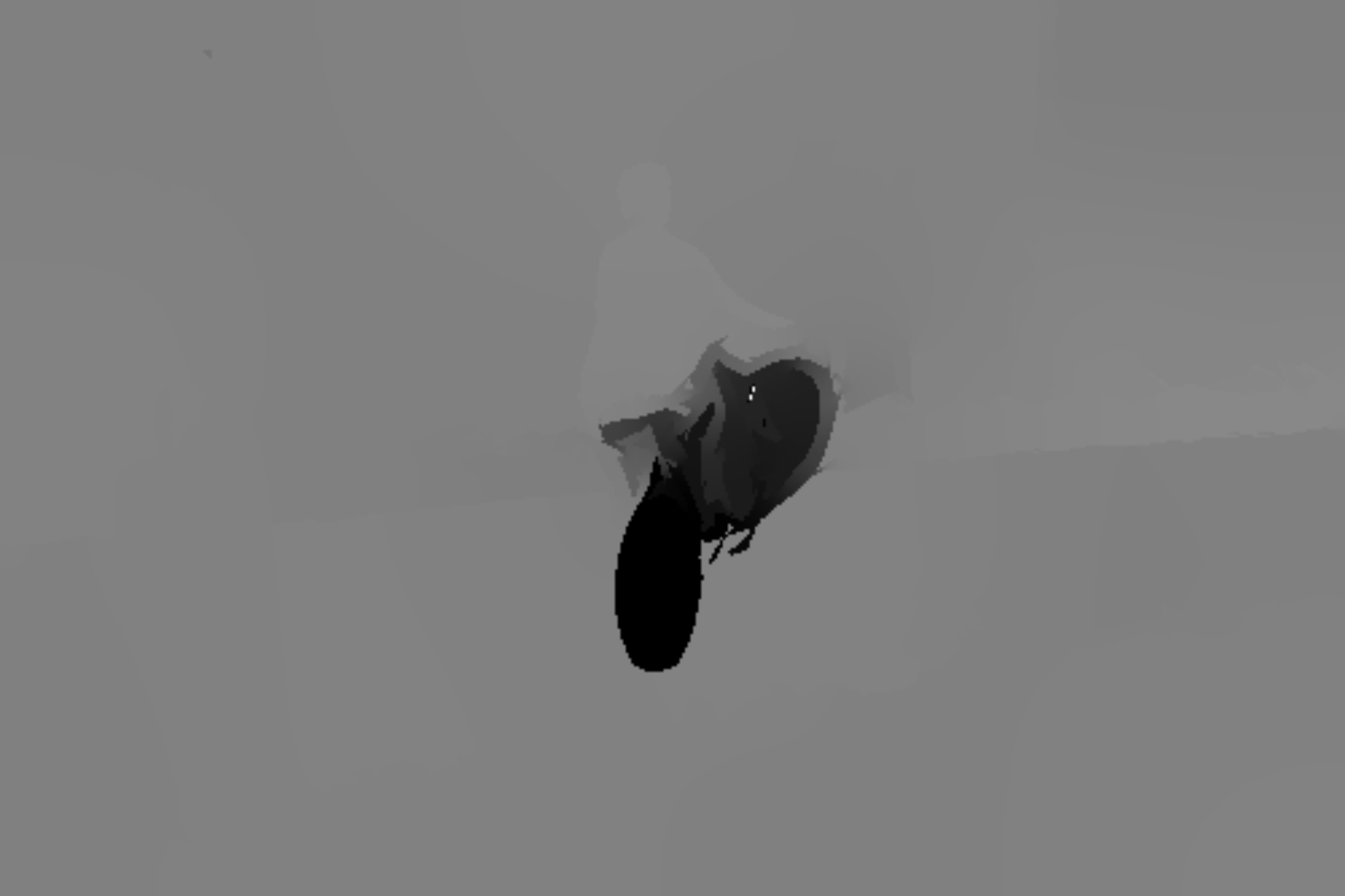}

\captionsetup{labelformat=default}
\setcounter{figure}{7}
    \caption{In this figure, the first column depicts an input image while the second and third columns illustrate two selected eigenvectors for that image. The eigenvectors contain soft segmentation information. Because \HfL boundaries capture object-level boundaries, the resulting eigenvectors primarily segment regions corresponding to the objects.\vspace{-0.3cm}}
    \label{eigv}
\end{figure}

\captionsetup{labelformat=default}

In this section, we present semantic segmentation results on the SBD~\cite{BharathICCV2011} and also Pascal VOC 2007~\cite{pascal-voc-2007} datasets, which both provide ground truth segmentations for $20$ Pascal VOC classes. We evaluate the results in terms of two metrics. The first metric measures the accuracy in terms of pixel intersection-over-union averaged per pixel (PP-IOU) across the 20 classes. According to this metric, the accuracy is computed on a per pixel basis. As a result, the images that contain large object regions are given more importance. 

We observe that while DeepLab-CRF works well on the images containing large object regions, it produces spatially disjoint outputs for the images with smaller and object regions (see Figure~\ref{qual_ss}).  This issue is often being overlooked, because according to the PP-IOU metric, the images with large object regions are given more importance and thus contribute more to the accuracy. However, certain applications may require accurate segmentation of small objects. Therefore, in addition to PP-IOU, we also consider the PI-IOU metric (pixel intersection-over-union averaged per image across the 20 classes), which gives equal weight to each of the images.


For both of the metrics we compare the semantic segmentation results of a pure DeepLab-CRF~\cite{DBLP:journals/corr/ChenPKMY14} and also a modification of DeepLab-CRF with our proposed features added to the CRF framework. We present the results for both of the metrics in Table~\ref{pp_iou}.

Based on these results, we observe that according to the first metric (PP-IOU), our proposed features yield almost equivalent results as the original DeepLab-CRF system. However, according to the second metric (PI-IOU) our features yield an average improvement of $3\%$ and $1.9\%$ in SBD and VOC 2007 datasets respectively. 


We also visualize the qualitative results produced by both approaches in Figure~\ref{qual_ss}. Notice how our proposed features make the segmentations look smoother relative to the segmentations produced by the original DeepLab-CRF system.

Once again, we want to stress that our \HfL  features are applicable to any method that uses the CRF. Therefore, based on the results presented in this section, we believe that our proposed features could be beneficial in a wide array of problems that involve the use of the CRF framework.

\captionsetup{labelformat=default}


%
%
%
%
%
%
%
%

\subsection{Object Proposals}
\label{obj_prop_gen}

Finally, we show that our method produces object-level boundaries that can be successfully exploited in an object proposal scheme. Specifically we adopt the EdgeBoxes approach~\cite{ZitnickDollarECCV14edgeBoxes}, which can be applied to any generic boundaries to produce a list of object proposal boxes. The original EdgeBoxes method uses SE boundaries to generate the boxes. However, SE boundaries are predicted using low-level color and texture features, rather than object-level features. Thus, here we validate the hypothesis that the EdgeBoxes proposals can be improved by replacing the SE boundaries with our \HfL boundaries.




\captionsetup{labelformat=empty}

\begin{figure}
\centering

\myfigurethreecol{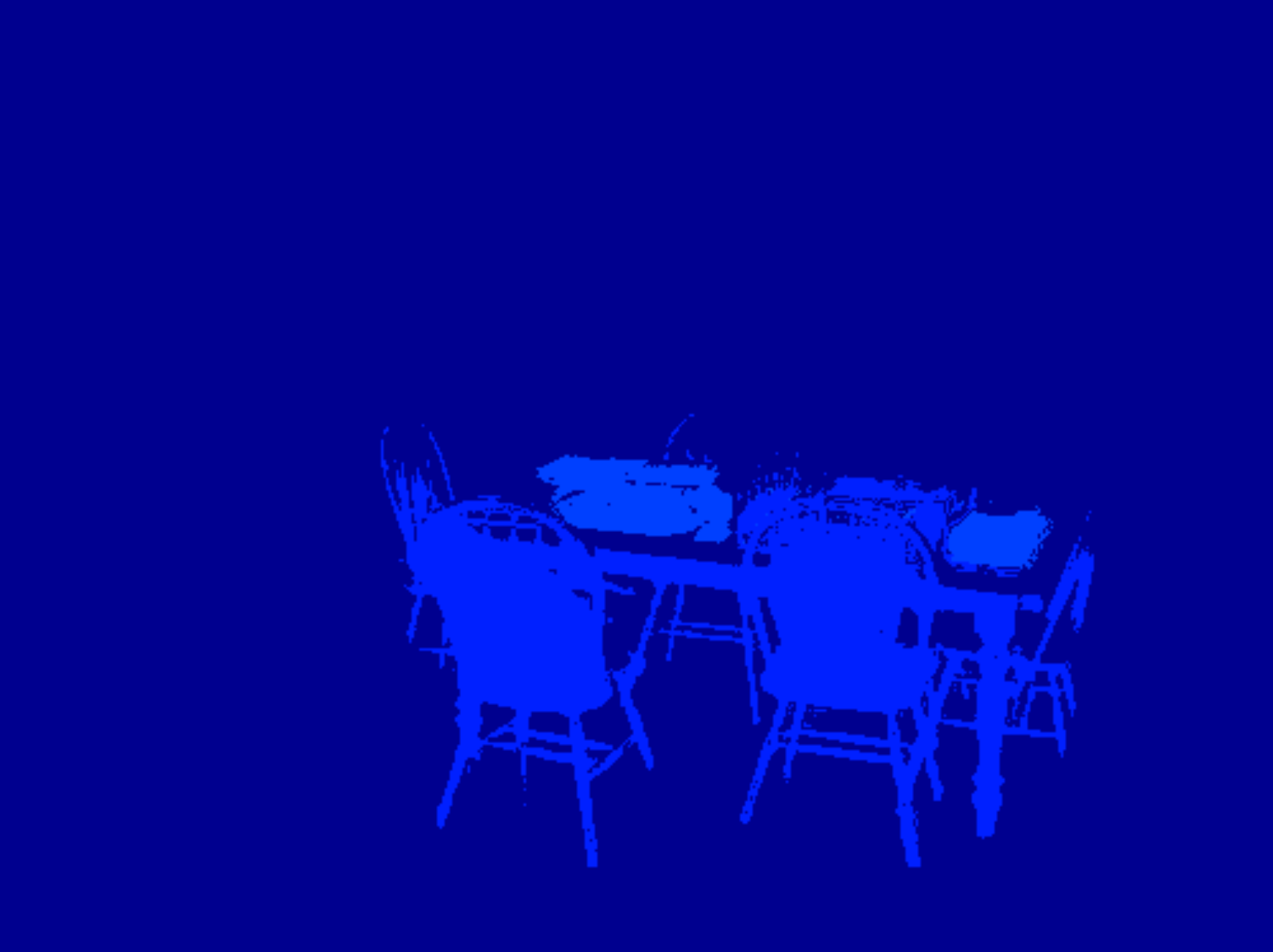}
\myfigurethreecol{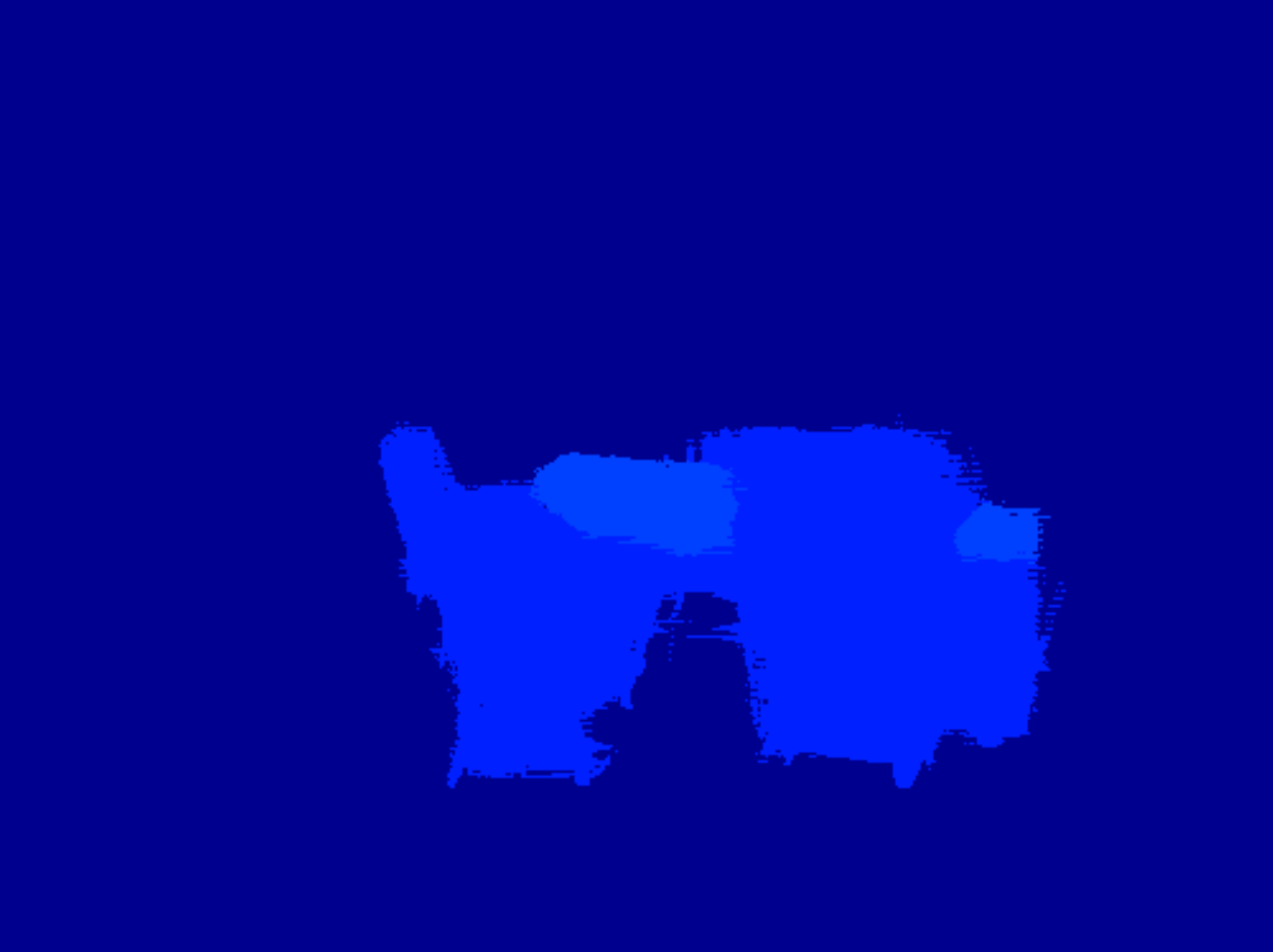}
\myfigurethreecol{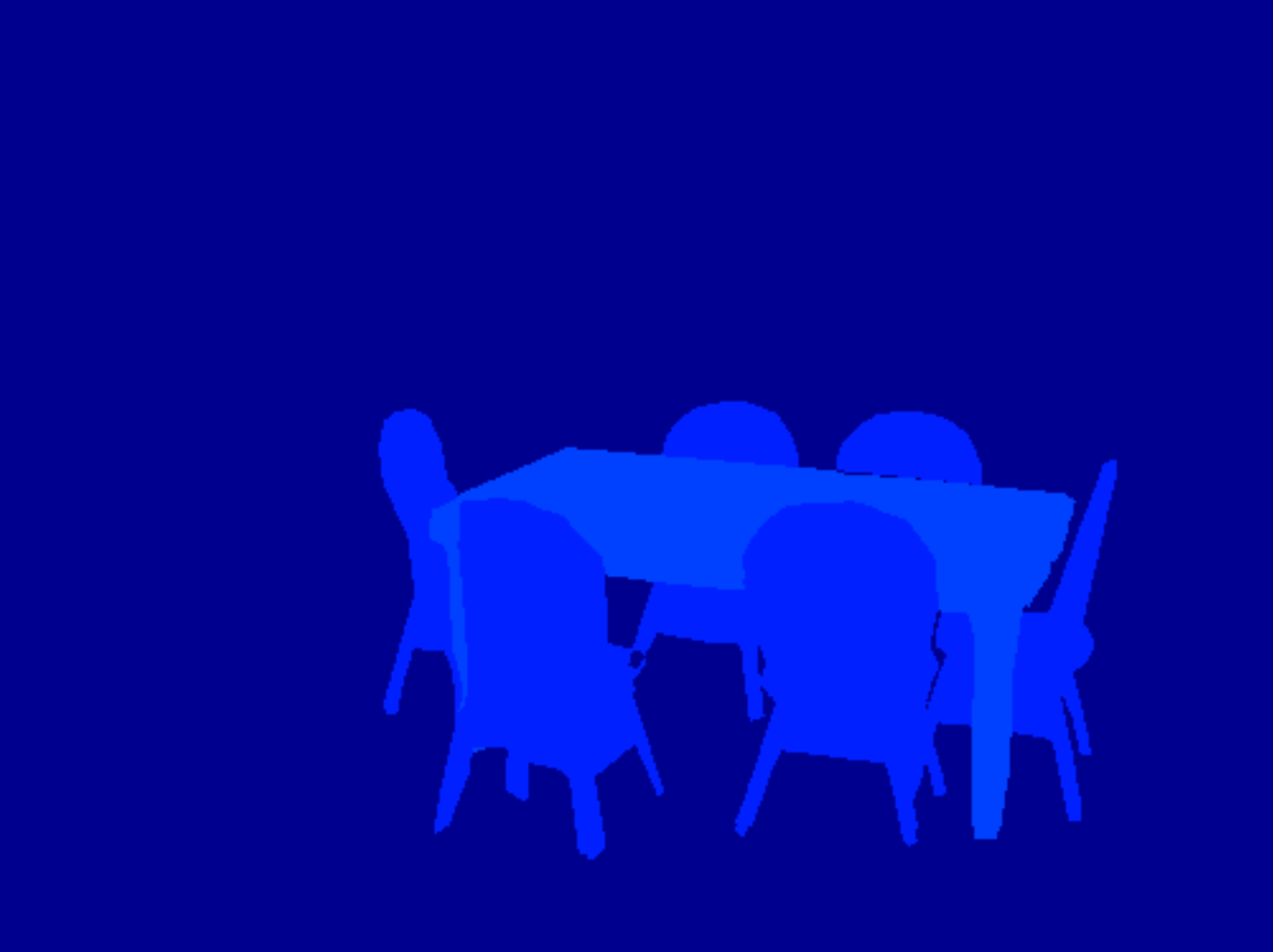}


\myfigurethreecol{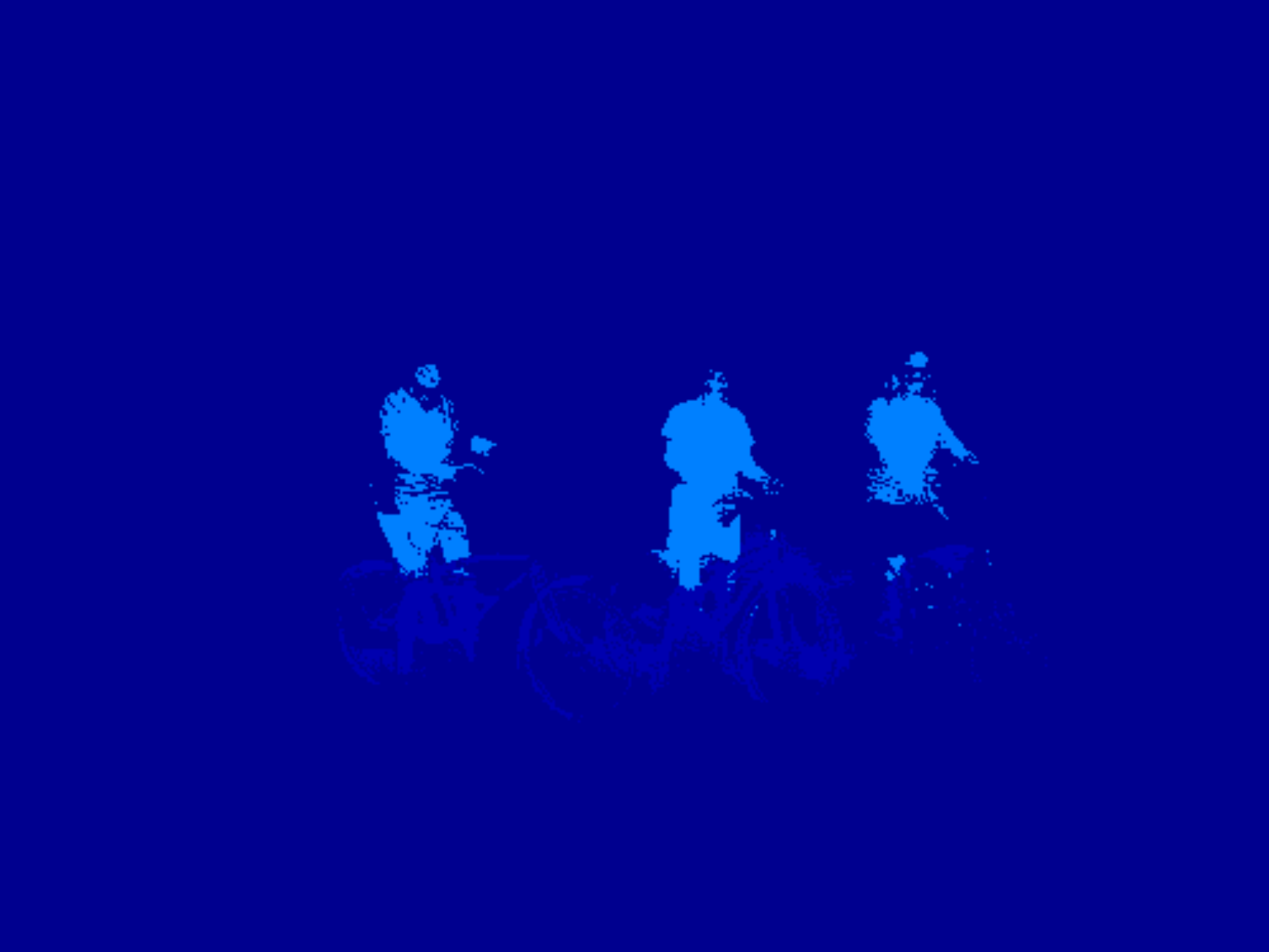}
\myfigurethreecol{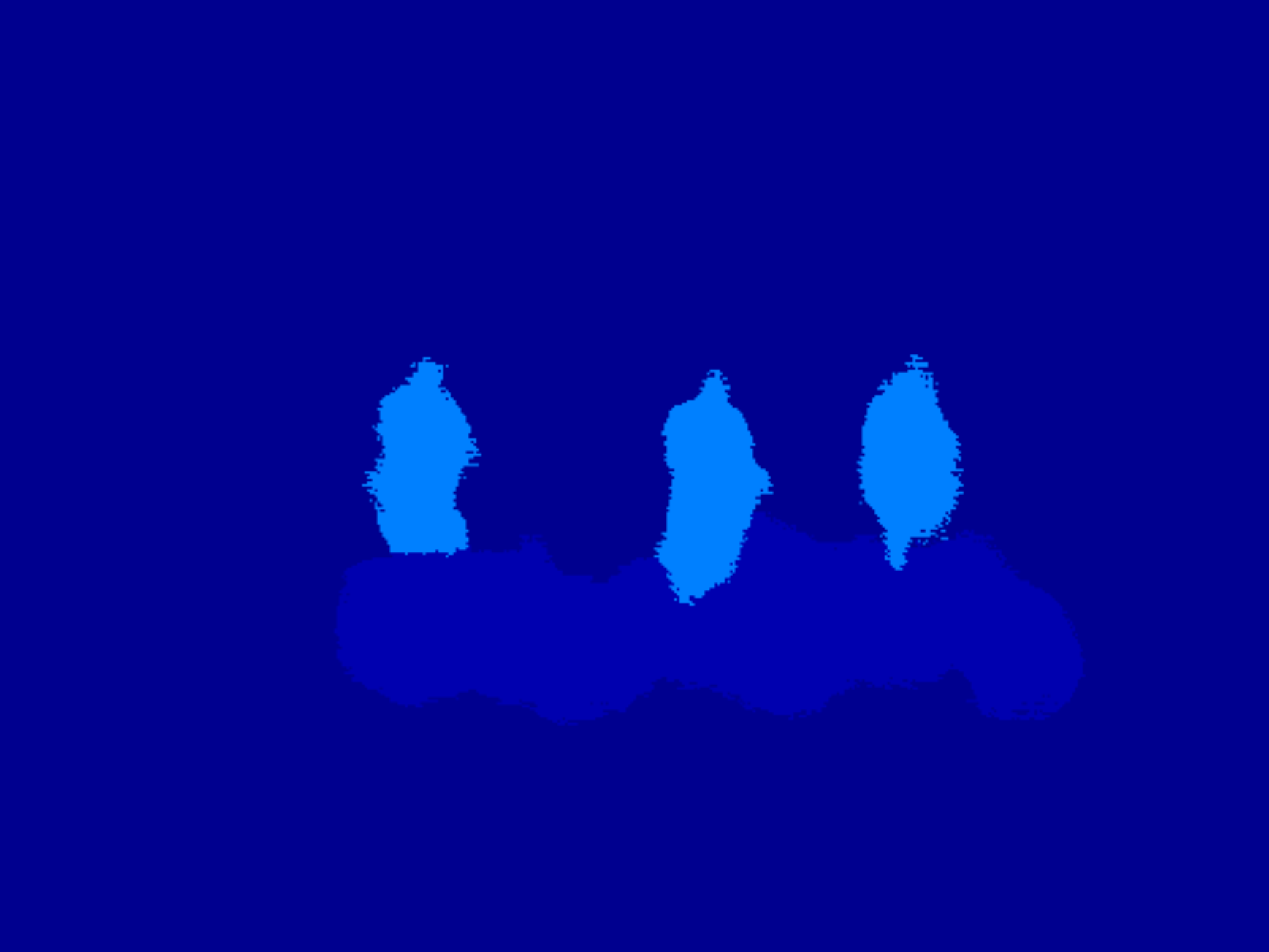}
\myfigurethreecol{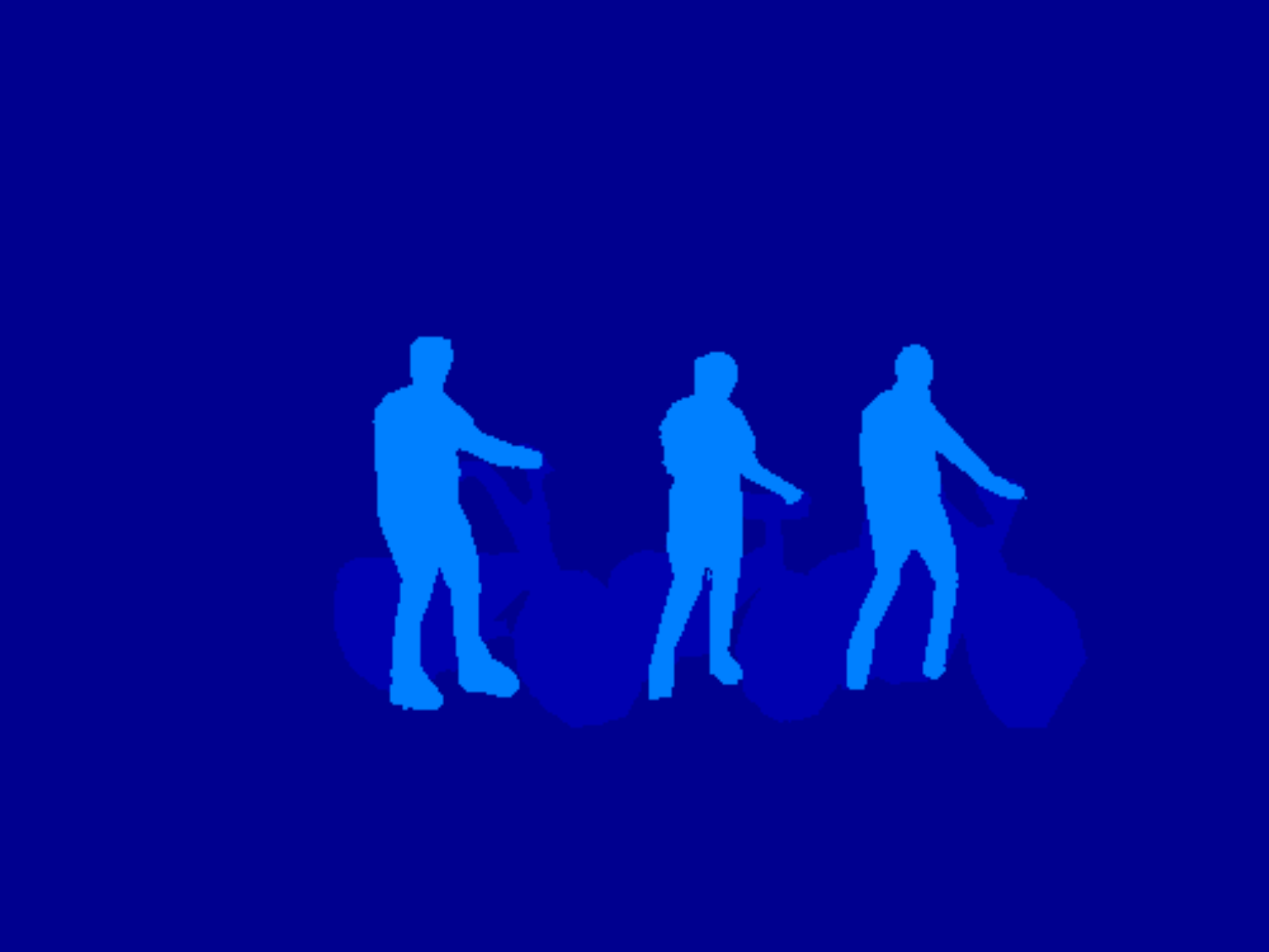}


\captionsetup{labelformat=default}
\setcounter{figure}{8}
    \caption{An illustration of the more challenging semantic segmentation examples. The first column depicts the predictions achieved by DeepLab-CRF, while the second column illustrates the results after adding our proposed features  to the CRF framework. The last column represents ground truth segmentations. Notice how our proposed features render the predicted semantic segments more spatially coherent and overall more accurate.}
    \label{qual_ss}
\end{figure}

\subsubsection{Object Proposal Results}

In this section, we present object proposal results on the Pascal VOC 2012 dataset~\cite{pascal-voc-2012}. We evaluate the quality of bounding-box proposals according to three metrics: area under the curve (AUC), the number of proposals needed to reach recall of $75\%$, and the maximum recall over $5000$ object bounding-boxes. Additionally, we compute the accuracy for each of the metrics for three different intersection over union (IOU) values: $0.65, 0.7$, and $0.75$. We present these results in Table~\ref{bb_results}. As described in Section~\ref{obj_prop_gen}, we use EdgeBoxes~\cite{ZitnickDollarECCV14edgeBoxes}, a package that uses generic boundaries, to generate object proposals. We compare the quality of the generated object proposals when using SE boundaries and \HfL boundaries. We demonstrate that for each IOU value and for each of the three evaluation metrics, \HfL boundaries produce better or equivalent results. This confirms our hypothesis that \HfL boundaries can be used effectively for high-level vision tasks such as generating object proposals.

\captionsetup{labelformat=default}

   \begin{table}
   \scriptsize
   \begin{center}
  \begin{tabular}{ l | SSS | SSS | SSS }
    \toprule
    \multirow{2}{*}{Method} &
      \multicolumn{3}{c |}{IoU 0.65} &
      \multicolumn{3}{c |}{IoU 0.7} &
      \multicolumn{3}{c }{IoU 0.75} \\
      & AUC & \text{N\text{@}75\%} & \text{Recall} & AUC & \text{N\text{@}75\%} & \text{Recall} & AUC & \text{N\text{@}75\%} & \text{Recall} \\
      \midrule
    SE & 0.52 & \hspace{0.25cm}413 & \hspace{0.1cm}0.93 & 0.47 & \hspace{0.2cm}658 & \hspace{-0.04cm}0.88 & \hspace{0.1cm}\bf 0.41 & \hspace{0.3cm}\text{inf} & 0.75 \\
    \bf \HfL & \hspace{0.17cm}\bf 0.53 & \hspace{0.25cm}\bf 365 & \hspace{0.1cm}\bf 0.95 & \hspace{0.16cm}\bf 0.48 & \hspace{0.2cm}\bf 583 & \hspace{0.05cm}\bf 0.9 & \hspace{0.1cm}\bf 0.41 & \hspace{0.2cm}\bf 2685 & \hspace{0.2cm}\bf 0.77 \\
    \bottomrule
  \end{tabular}
  \end{center}
      \caption{Comparison of object proposal results. We compare the quality of object proposals using Structured Edges~\cite{Dollar2015PAMI}  and \HfL boundaries. We evaluate the performance for three different IOU values and demonstrate that using  \HfL boundaries produces better results for each evaluation metric and for each IOU value.}
    \label{bb_results}
\end{table}

\section{Conclusions}

In this work, we presented an efficient architecture that uses object-level information to predict semantically meaningful boundaries. Most prior edge detection methods rely exclusively on low-level features, such as color or texture, to detect the boundaries. However, perception studies suggest that humans employ object-level reasoning when deciding whether a given pixel is a boundary~\cite{psych,sanguinetti2013ground,KourtziKanwisher01}. Thus, we propose a system that focuses on the semantic object-level cues rather than low level image information to detect the boundaries. For this reason we refer to our boundary detection scheme as a \textit{High-for-Low} approach, where high-level object features inform the low-level boundary detection process. In this paper we demonstrated that our proposed method produces boundaries that accurately separate objects and the background in the image and also achieve higher F-score compared to any prior work.


Additionally, we showed that, because \HfL boundaries are based on object-level features, they can be employed to aid a number of high level vision tasks in a \textit{Low-for-High} fashion. We use our boundaries to boost the accuracy of state-of-the-art methods on the high-level vision tasks of semantic boundary labeling, semantic segmentation, and object proposals generation. We show that using \HfL boundaries leads to better results in each of these tasks.

To conclude, our boundary detection method is accurate, efficient, applicable to a variety of datasets, and also useful for multiple high-level vision tasks. We plan to release the source code for \HfL upon the publication of the paper .

\section{Acknowledgements}

We thank Mohammad Haris Baig for the suggestions and help with the software. This research was funded in part by NSF award CNS-1205521.

{\small
\bibliographystyle{ieee}
\bibliography{gb_bibliography}
}

\end{document}